\RequirePackage{fix-cm}
\documentclass[11pt, a4paper, copyright, gr]{google}

\usepackage[authoryear, sort&compress, round]{natbib}
\usepackage{graphicx}
\usepackage{booktabs}
\usepackage{multirow}
\usepackage{amsmath,amssymb,amsfonts}
\usepackage{amsthm}
\usepackage{caption}
\usepackage{xcolor}
\usepackage{textcomp}
\usepackage{enumitem}
\usepackage{longtable}
\usepackage{float}
\usepackage{array}
\usepackage{xurl}

\graphicspath{{./}{figures/}}

\providecommand{\botrule}{\bottomrule}

\title{Planetary Geospatial Foundation Models: A New Paradigm for Global Public Health}

\reportnumber{} 

\correspondingauthor{\space\unskip\{shekelto, arbaazmuslim, sshetty, gautamprasad\}@google.com}

\uselogo{}

\author[*,1]{Arbaaz Muslim}
\author[*,1]{Aviv Slobodkin}
\author[*,2]{Katherine Wheeler-Martin}
\author[*,3,4,5,6]{Eric Zhou}
\author[*,7,8]{John Brittain}
\author[*,9,10,11]{Martin Frasch}

\author[1]{Hamsa Subramaniam}
\author[1]{Jacob Bien}
\author[1]{Hamed Sadeghi}
\author[2]{Sarah Conderino}
\author[1]{Stone Jiang}
\author[2]{Ben Spoer}
\author[2]{Daniel Neill}
\author[1]{Mimi Sun}
\author[1]{Joydeep Paul}
\author[1]{Yun Liu}
\author[1]{Tomer Shekel}
\author[1]{Amy Chung-Yu Chou}
\author[1]{Anna Carter}
\author[1]{Charles Elliott}
\author[1]{Aaron Bell}
\author[1]{Reuven Sayag}
\author[1]{Avinatan Hassidim}
\author[1]{Niv Efron}
\author[1]{Yossi Matias}
\author[1]{Kasumi Widner}
\author[12]{Eduardo Lopez Ortiz}
\author[13]{Jos\'e Alberto D\'{i}az-Qui\~nonez}
\author[7,8]{Moritz U. G. Kraemer}
\author[6,14]{John Brownstein}
\author[15]{Isaias Fernandes Co}

\author[$\dag$,16]{Etien Luc Koua}
\author[$\dag$,1]{Monica Bharel}
\author[$\dag$,6,17]{Benjamin Rader}
\author[$\dag$,2]{Lorna Thorpe}
\author[$\dag$,2]{Marc Gourevitch}
\author[$\dag$,1]{Shravya Shetty}
\author[$\dag$,1]{Gautam Prasad}

\affil[*]{Equal contribution}
\affil[$\dag$]{Co-last author}
\affil[1]{Google, Mountain View, CA, USA}
\affil[2]{Department of Population Health, NYU Grossman School of Medicine, New York, NY, USA}
\affil[3]{Center for Child Health Services Research, Icahn School of Medicine at Mount Sinai, New York, NY, USA}
\affil[4]{Department of Pediatrics, Icahn School of Medicine at Mount Sinai, New York, NY, USA}
\affil[5]{Department of Population Health Science and Policy, Icahn School of Medicine at Mount Sinai, New York, NY, USA}
\affil[6]{Computational Epidemiology Lab, Boston Children's Hospital, Boston, MA, USA}
\affil[7]{Department of Biology, University of Oxford, Oxford, UK}
\affil[8]{Pandemic Sciences Institute, University of Oxford, Oxford, UK}
\affil[9]{Institute on Human Development and Disability, University of Washington, Seattle, WA, USA}
\affil[10]{Health Stream Analytics, LLC, Seattle, WA, USA}
\affil[11]{KHAI Ventures, Boston, MA, USA}
\affil[12]{National Autonomous University of Mexico, Mexico City, Mexico}
\affil[13]{Tecnol\'ogico de Monterrey, School of Medicine and Health Sciences, Mexico City, Mexico}
\affil[14]{Departments of Pediatrics and Biomedical Informatics, Harvard Medical School, Boston, MA, USA}
\affil[15]{World Health Organization Regional Office for Africa (WHO AFRO), Nairobi, Kenya}
\affil[16]{Emergency Preparedness and Response, World Health Organization Regional Office for Africa (WHO AFRO), Brazzaville, Republic of the Congo}
\affil[17]{Department of Anaesthesia, Harvard Medical School, Boston, MA, USA}

\begin{abstract}
The efficacy of traditional disease prediction is limited by spatial gaps and temporal lags, which impact the timing and targets of resource deployments. Outbreaks escalate undetected~\citep{bib1}, chronic disease burdens are quantified years later~\citep{bib2}, and at-risk populations in data-sparse regions remain unaddressed~\citep{bib3}. Planetary geospatial foundation models~\citep{bib4} complement existing epidemiological workflows to provide operational improvements, encoding multimodal search, mobility, and environmental signals into generalizable place representations. As illustrations of this complementarity, we present independent global health case studies of Google Earth AI's Population Dynamics Foundation Model (PDFM) -- a foundation model for geospatial inference~\citep{bib4} -- across four domains (vaccine-preventable, communicable, noncommunicable, maternal mental health), five tasks (spatial extrapolation, interpolation/nowcasting, probabilistic forecasting, prospective forecasting, risk stratification), and four countries (USA, Canada, Mexico, and the Democratic Republic of the Congo). Across these case studies, PDFM addresses critical surveillance gaps across domains: improving US--Canada border MMR vaccination coverage predictions by capturing cross-border behavioral spillovers domestic models miss; nowcasting cardiovascular disease to accelerate data availability; enhancing short-term municipal Mexican dengue forecasts for timely outbreak vector control; improving forecasts of cholera hotspots; and adding a transferable signal to individual-level postpartum-depression risk prediction in US states the model had never seen, while not replacing individual socioeconomic data or closing demographic screening gaps. Together, these results showcase capabilities of geospatial foundation models for public health surveillance.
\end{abstract}

\begin{document}

\maketitle

\section{Introduction}

Global public health operates at a critical crossroads: overlapping infectious disease outbreaks, a rising burden of noncommunicable diseases, persistent workforce shortages, systemic health disparities, and chronic underfunding collectively threaten population health outcomes worldwide, with far-reaching economic and societal consequences. In this landscape, recent advances in artificial intelligence offer tools to strengthen disease prediction and response by better characterizing the factors that drive health outcomes, identifying populations most at risk, and directing resources when and where they are most needed. Yet this promise hinges on data availability, and conventional epidemiological data streams carry systemic limitations that constrain modern prediction models. Over time, demographic and socioeconomic data from censuses and surveys, typically updated annually and rarely comprehensive, struggle to capture rapid behavioral and environmental shifts across geospatial parameters. Obtaining disease outcome data at the fine spatial resolutions needed to guide interventions, such as local mortality counts, is often impossible in real time. Nearly all conventional data sources stop at national boundaries, missing the cross-border dynamics that shape health outcomes in an interconnected world~\citep{bib5,bib6}.

These data gaps directly compromise public health tracking and disease prediction. For example, country-level vaccination surveillance and modeling stop at national borders, missing the cross-border mobility, economic integration, and shared media environments that shape immunization behavior across the approximately 18.6 million people living in US counties within 150 km of the Canadian border---blinding local risk assessments at a time when national kindergarten MMR coverage has fallen to 92.7\% and measles outbreaks have surged~\citep{bib7}. Additionally, unsuppressed county-level cardiovascular disease (CVD) mortality data from the National Vital Statistics System typically lags 1--2 years behind real time~\citep{bib8}, forcing health departments to allocate prevention resources based on outdated information for a disease that killed 916,000 Americans in 2023~\citep{bib9}, corresponding to nearly three in ten fatalities. These challenges are further exacerbated by the resource-intensive nature of conventionally task-specific disease prediction: each new disease, geography, or forecasting horizon demands dedicated model training and labeled ground-truth datasets, limiting scalability when rapid response is most needed.

Against these limitations, a recent class of planetary geospatial foundation models offers a fundamentally different approach to empower public health surveillance beyond the reach of conventional data. These models draw on privacy-preserving, multimodal data streams spanning population behavior including aggregated web search trends, human mobility patterns, built-environment characteristics, meteorological conditions, and air quality measurements. Through self-supervised pre-training at the planet level---learning patterns from the structure of these signals without requiring disease-specific outcome labels---they produce rich, transferable geospatial representations that encode how populations interact with their environments and with each other, generalizing across diseases, geographies, and modeling tasks. Driven by the rapid evolution of these models, platforms such as Google's Earth AI geospatial foundation program~\citep{bib10} have emerged to scale their insights globally. Within this suite, the Population Dynamics Foundation Model (PDFM)~\citep{bib4} serves as a proof-of-concept, producing geospatial feature vectors that can be seamlessly integrated as covariates into the diverse statistical and machine learning tools already used by public health practitioners---complementing rather than replacing established data sources and methods.

Here we evaluate whether PDFM embeddings can meaningfully improve disease prediction through independent evaluations conducted by multiple global health research partners across a broad range of public health tasks. To establish that these improvements hold broadly across the core dimensions that shape global health surveillance---epidemiological task, geographic setting, and disease type---we structure these evaluations along three complementary axes. The first varies the epidemiological task: cross-border extrapolation of vaccination trends, where PDFM captures behavioral spillover effects that domestic-only models miss; spatial interpolation and temporal nowcasting of cardiovascular disease mortality, where PDFM-derived covariates serve as a timely updating substitute, refreshing at a monthly cadence, for census data that grows stale between multi-year update cycles; short-term dengue outbreak forecasting, where PDFM corrects large prediction errors in active outbreak hotspots; prospective outbreak forecasting in low-resource contexts, where PDFM supplements historical data to provide valuable 4- to 8-week lead time in cholera hotspot prediction; and individual-level risk prediction for postpartum depression, where PDFM adds a signal in unseen states but does not replace individual socioeconomic data or reduce demographic biases, marking a boundary of the approach. The second spans geographic and resource context, from high-resource settings with rich data infrastructure (United States, Canada) to data-scarce environments with active humanitarian crises (DRC). The third varies the disease type, covering both communicable diseases (dengue in Mexico and cholera in the DRC) and noncommunicable conditions (cardiovascular disease mortality and postpartum depression in the United States). By synthesizing these partner-led evaluations thematically rather than as isolated case studies, we identify broadly applicable insights into the capabilities and boundaries of geospatial foundation models for public health surveillance.

\section{Results}

\begin{figure}[htbp]
\centering
\includegraphics[width=\textwidth]{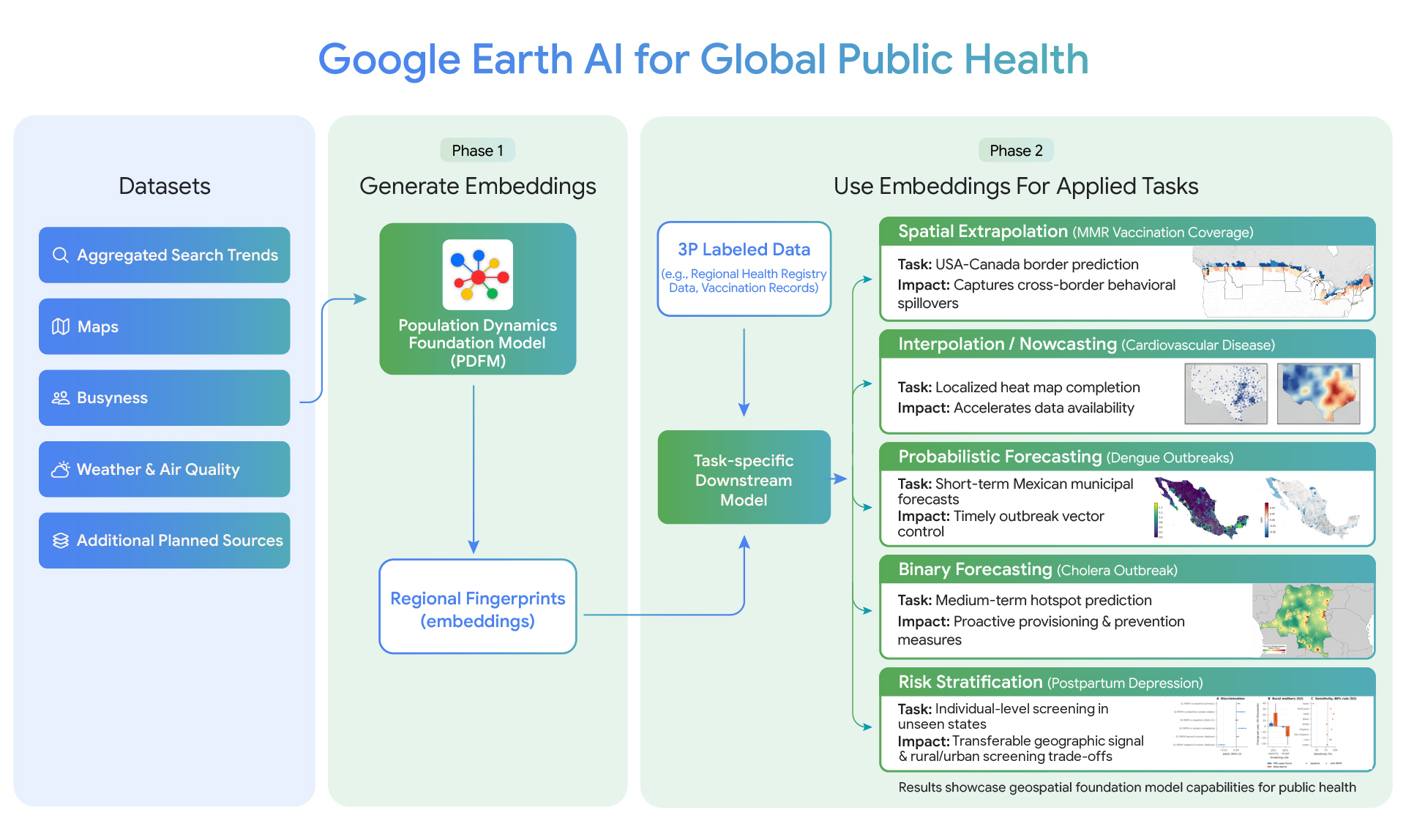}
\caption{Overview of Google Earth AI for global public health. Phase 1 encodes multimodal, privacy-preserving data streams (aggregated search trends, maps, busyness, and weather and air quality) via the Population Dynamics Foundation Model (PDFM) into generalizable regional fingerprint embeddings. Phase 2 integrates these place representations with domain-specific labeled health datasets and downstream statistical or machine-learning models across five epidemiological tasks: spatial extrapolation (MMR vaccination coverage), spatial interpolation and nowcasting (cardiovascular disease mortality), probabilistic forecasting (dengue outbreaks), prospective binary forecasting (cholera emergence), and individual-level risk stratification (postpartum depression).}\label{fig1}
\end{figure}

We evaluate the utility of geospatial foundation model embeddings across five case studies organized around three core epidemiological roles: (1) overcoming spatial borders and multi-year census reporting lags in population surveillance (Case Studies 1--2); (2) extending outbreak lead times across the critical 1-to-2-month operational window in infectious disease hotspots (Case Studies 3--4); and (3) establishing the boundary between place-level context and individual-level clinical risk stratification (Case Study 5). Case Study 1 tests cross-border extrapolation, assessing whether contextual embeddings from Canadian postal areas improve prediction of measles--mumps--rubella vaccination uptake in nearby U.S. counties. Case Study 2 compares these embeddings against standard socioeconomic covariates for interpolating and nowcasting county-level cardiovascular disease mortality. Case Study 3 evaluates probabilistic dengue forecasting across approximately 2,450 Mexican municipalities at multiple time horizons, measuring whether embeddings improve forecast accuracy and how that improvement varies by local disease burden. Case Study 4 tests whether adding custom 200-dimensional PDFM embeddings to historical surveillance features improves prospective forecasts of cholera emergence, 1 to 8 weeks ahead, across 403 health zones in the DRC. Finally, Case Study 5 tests whether geographic context adds to individual-level postpartum-depression risk prediction, whether it can substitute for individual socioeconomic data, and whether it narrows demographic performance gaps, without using race or ethnicity as model inputs. In all cases, the embeddings are used as fixed covariates without task-specific fine-tuning, combined with domain-appropriate statistical or machine-learning models to assess the foundation model's out-of-the-box representational value. Together, these evaluations span multiple modeling paradigms, geographies, and disease domains, providing a unified assessment of both where geospatial foundation models deliver operational value and where their boundaries lie (see Figure~\ref{fig1} and Table~\ref{tab1}).

\begin{table}[htbp]
\caption{Summary of Evaluation Benchmarks and Dataset Specifications. Overview of target variables, geographic scopes, and spatial resolutions across the five evaluation paradigms.}\label{tab1}%
\fontsize{7.5}{9}\selectfont
\setlength{\tabcolsep}{0pt}
\begin{tabular*}{\textwidth}{@{\extracolsep\fill}>{ \raggedright\arraybackslash }p{0.12\textwidth}>{ \centering\arraybackslash }p{0.11\textwidth}>{ \centering\arraybackslash }p{0.14\textwidth}>{ \centering\arraybackslash }p{0.13\textwidth}>{ \centering\arraybackslash }p{0.14\textwidth}>{ \centering\arraybackslash }p{0.14\textwidth}>{ \centering\arraybackslash }p{0.15\textwidth}@{}}
\toprule
\textbf{ML Problem} & \textbf{Geographic Scope} & \textbf{Thematic Domain} & \textbf{Benchmark Dataset} & \textbf{Prediction Targets} & \textbf{Train / Test Granularity} & \textbf{Train / Test Size} \\
\midrule
Spatial Extrapolation & Global North (US, Canada) & Vaccine-Preventable (Immunization) & MMR Vaccination Uptake (ONM/MRP) & County-level MMR vaccination rate & US County / Canadian FSA (postal) & 3,080 US counties (10-fold cross-validation [CV]); 146 border counties subset \\
\addlinespace[5pt]
Interpolation & Global North (US) & Non-communicable (Cardiovascular) & NVSS CVD Mortality & County-level CVD death counts & US County (Census-equivalent) & 3,091 counties (5-fold CV) \\
\addlinespace[5pt]
Nowcasting & Global North (US) & Non-communicable (Cardiovascular) & NVSS CVD Mortality & County-level CVD death counts & US County (Census-equivalent) & 30,910 county-years (2013--2022) / 3,091 counties (2023) \\
\addlinespace[5pt]
Probabilistic Forecasting & Global South (Mexico) & Communicable (Dengue) & Mexico Dengue Surveillance for the Mexican Ministry of Health & Monthly dengue case counts & GADM Admin Level 2 (municipality) & $\sim$2,450 municipalities $\times$ 68 months (Jan 2020--Aug 2025); 24-month sliding window \\
\addlinespace[5pt]
Prospective Forecasting & Global South (DRC) & Communicable (Cholera) & DRC National Integrated Disease Surveillance & Binary cholera emergence at 1-, 2-, 4-, and 8-week leads & Health zone $\times$ week & 403 health zones; train: 722 weeks (4 Jan 2010--31 Oct 2023) / test: 89 weeks (1 Nov 2023--14 Jul 2025) \\
\addlinespace[5pt]
Risk Stratification & Global North (US) & Maternal Mental Health (PPD) & CDC PRAMS, Phases 7--8 (births 2012--2021) & Binary postpartum depressive symptoms (CDC indicator: MH\_PPDPR or MH\_PPINT ``always'' or ``often'') & Respondent; PDFM v0 embedding at state $\times$ urban/rural (87 locations) & 332,970 respondents; sealed tests: 66,594 holdout (S1), 10 jurisdiction-held-out folds (S2), 2020--21 births (S3) \\
\botrule
\end{tabular*}
\end{table}

\subsection{Case Study 1: Cross-border geospatial embeddings capture behavioral spillovers to improve localized immunization forecasting}

Vaccination coverage near international borders is not necessarily governed by political boundaries alone: shared media environments, population mobility, and economic integration may create cross-border behavioral patterns that purely domestic models fail to capture. This case study tests whether incorporating PDFM embeddings from Canadian postal areas as contextual features can improve prediction of measles--mumps--rubella (MMR) vaccination uptake in U.S. counties along the U.S.--Canada border.

To incorporate cross-border context, we aggregated PDFM embeddings from Canadian postal areas within a 150 km radius of each U.S. county using inverse-distance weighted spatial kernels, giving greater weight to nearby areas~\citep{bib11}. We compared three model specifications: U.S. embeddings only (baseline), Canadian contextual embeddings only, and combined U.S. and Canadian embeddings. Performance was assessed by correlation, RMSE, MAE, and R$^2$ between predicted and observed vaccination rates, both nationally and among the subset of 146 border-adjacent counties.

At the national scale (see Table~\ref{tab2}, top), the U.S.-only baseline achieved a correlation of 0.611, RMSE of 0.058, and R$^2$ of 0.373, while using Canadian contextual embeddings alone performed poorly nationally (correlation 0.113, RMSE of 0.072, and R$^2$ = 0.013), confirming that cross-border information cannot substitute for domestic covariates. Interestingly, though modest on the national scale, we find that combining the two does have a positive effect across all metrics.

The value of cross-border context emerges more clearly when focusing specifically on the border counties (Table~\ref{tab2}, bottom). In this subset, the U.S.-only model and the Canadian contextual embeddings alone achieve comparable performance across all metrics, suggesting that the Canadian embeddings already capture meaningful local information within these border U.S. counties. Importantly, adding Canadian embeddings to the U.S. model yields significant gains in performance across all metrics, most notably in R$^2$, which increases from 0.159 to 0.216---a roughly 36\% relative increase in explained variance.

To characterize the practical implications of these gains, we examined how many residents live in border counties where adding Canadian contextual embeddings shifted predicted MMR coverage by a meaningful margin (Figure~\ref{fig3} and Extended Data Table~\ref{exttab1}). In 24 counties, encompassing approximately 2.5 million residents (13.4\% of the border-county population), predictions changed by at least 5 percentage points; the number of affected residents grew to 4.7 million (25.3\%) when considering shifts of at least 3 percentage points, and to 13.1 million (70.5\%) for shifts of at least 1 percentage point. These results indicate that incorporating cross-border contextual information materially alters estimated vaccination coverage across a substantial share of U.S. border residents.

\begin{table}[htbp]
\caption{Predictive performance of models using U.S. and Canadian contextual information to predict county-level MMR vaccination coverage. We show results for all 3,080 US counties (top three rows) and for all 146 counties on the US--Canada border (bottom three rows). For each of these test cases, the best results on each of the reported metrics are in bold.}\label{tab2}%
\setlength{\tabcolsep}{0pt}
\begin{tabular*}{\textwidth}{@{\extracolsep\fill}llcccc@{}}
\toprule
\textbf{Geography} & \textbf{Model} & \textbf{Correlation} & \textbf{RMSE} & \textbf{MAE} & \textbf{$R^2$} \\
\midrule
\multirow{3}{*}{All US counties} & US embeddings only & 0.611 & 0.058 & 0.045 & 0.373 \\
 & Canadian context only & 0.113 & 0.072 & 0.059 & 0.013 \\
 & US + Canadian context & \textbf{0.614} & \textbf{0.057} & \textbf{0.044} & \textbf{0.381} \\
\midrule
\multirow{3}{*}{US--CA border counties} & US embeddings only & 0.399 & 0.070 & 0.055 & 0.159 \\
 & Canadian context only & 0.381 & 0.071 & 0.056 & 0.146 \\
 & US + Canadian context & \textbf{0.465} & \textbf{0.068} & \textbf{0.051} & \textbf{0.216} \\
\botrule
\end{tabular*}
\end{table}

\begin{figure}[!htb]
\centering
\includegraphics[width=0.75\textwidth]{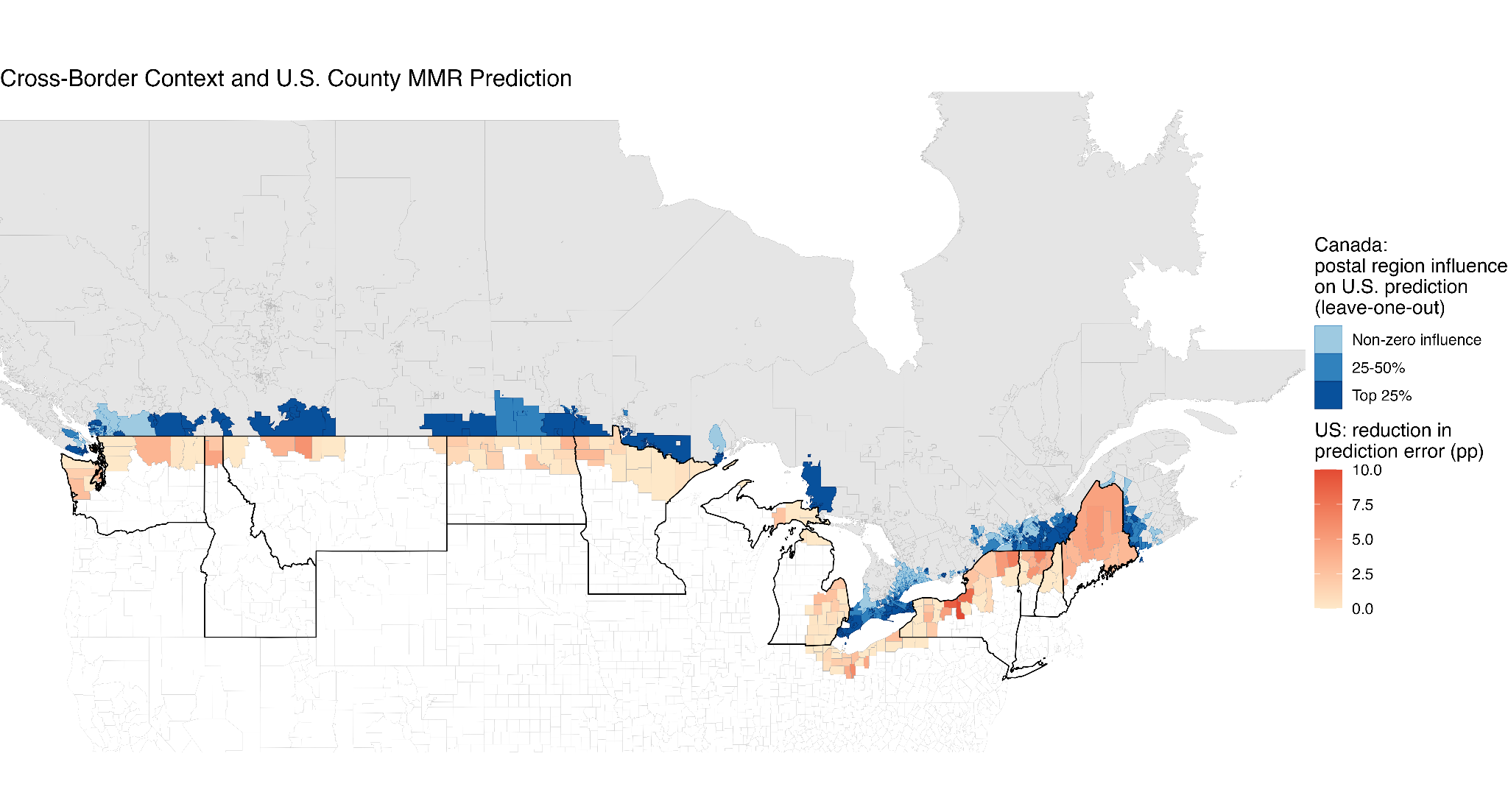}
\caption{Spatial distribution of cross-border predictive influence between Canadian postal regions and U.S. counties. The map illustrates where Canadian contextual embeddings contribute to improved prediction of U.S. county vaccination rates. Orange shading indicates U.S. counties where incorporating Canadian contextual embeddings reduces prediction error relative to a model using U.S. embeddings alone; darker shades correspond to larger reductions in absolute prediction error. Blue shading indicates Canadian postal regions whose embeddings exert the greatest influence in the cross-border prediction model, measured by the increase in prediction error that occurs when the information from that region is removed from the model through a leave-one-out procedure.}\label{fig2}
\end{figure}

\begin{figure}[!htb]
\centering
\includegraphics[width=0.58\textwidth]{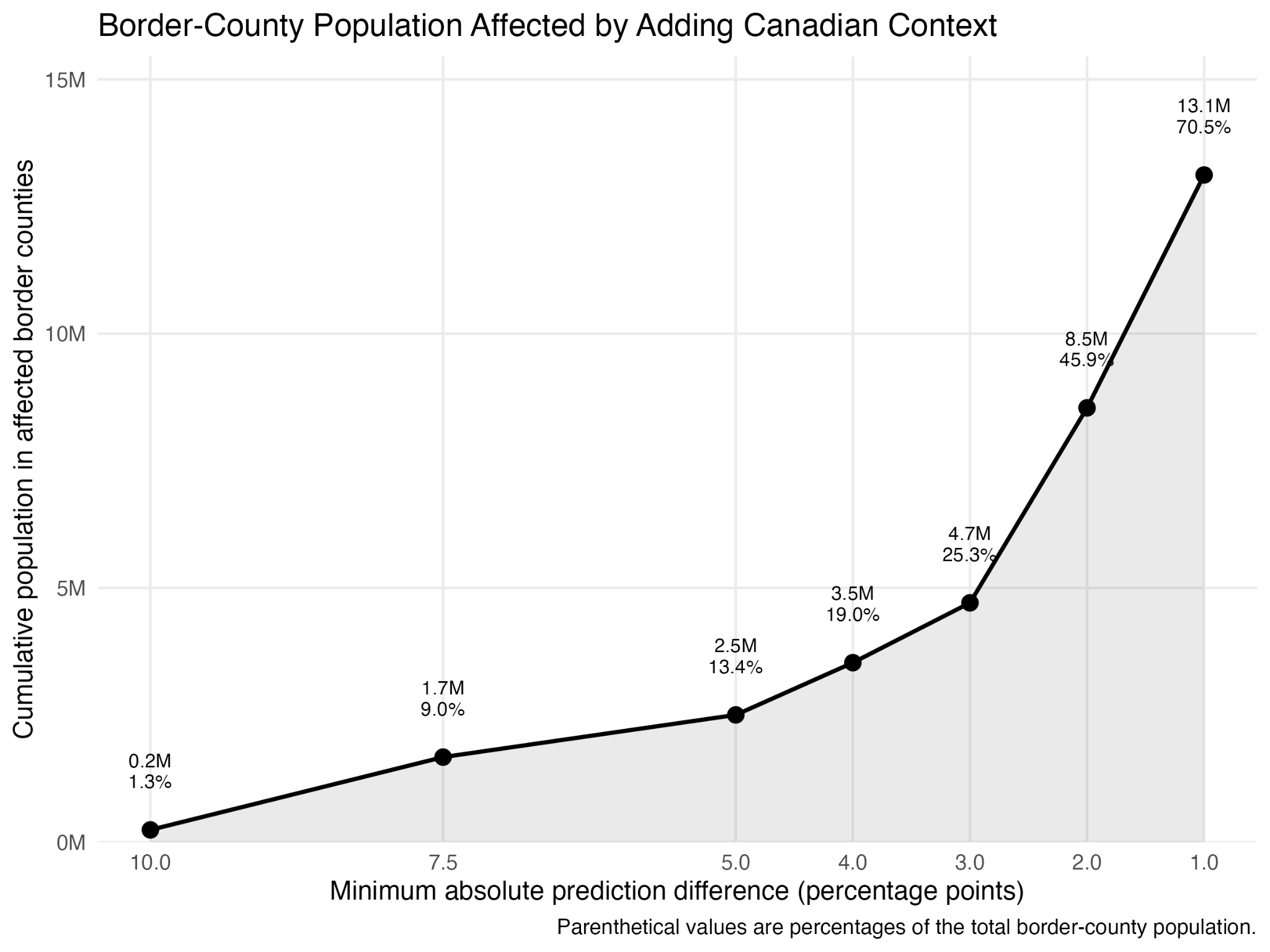}
\caption{Cumulative population living in U.S. border counties whose predicted vaccination coverage changes by at least a given threshold after incorporating Canadian contextual information. For each threshold, the corresponding affected population is presented as both population count and the percentage of the total border-county population.}\label{fig3}
\end{figure}

Figure~\ref{fig2} illustrates the geographic distribution of these effects, and identifies the Canadian postal regions whose embeddings contribute most strongly to the cross-border predictive signal. These regions are mainly concentrated in southern Ontario, southern Quebec, and southwestern British Columbia, which are areas naturally situated to have strong economic ties, population mobility, and media integration with adjacent U.S. counties~\citep{bib12}. The prediction gains provided by incorporating Canadian postal region data suggest that the embeddings encode localized cross-border contextual factors not captured by domestic predictors alone, and that spatial context extending beyond national boundaries can provide complementary information for modeling public health outcomes.

\subsection{Case Study 2: Population dynamics for cardiovascular disease nowcasting and interpolation}

Cardiovascular disease is a leading cause of death in the United States. However, proactive management of this crisis is hindered by limited data availability: unsuppressed mortality data typically lags by 1--2 years, while timely provisional data faces substantial suppression at fine spatiotemporal scales. These limitations hinder accurate allocation of preventative resources. Extending from cross-border spatial gaps (Case Study 1) to temporal reporting lags within a national surveillance system, this case study evaluates whether single-month PDFM embeddings can serve as a timely, high-frequency substitute or complement for multi-year census covariates when modeling county-level cardiovascular disease mortality.

Two tasks were evaluated: spatial interpolation, where mortality data are observed for some counties but not others at a single time point; and nowcasting, where historical data are available for all counties but present-day counts are only known for some counties. Performance was assessed by MAE and RMSE against held-out observations (Tables~\ref{tab3}--\ref{tab4}). For each task, two model classes were evaluated: negative binomial Bayesian spatiotemporal (Besag--York--Molli\'e) models and gradient-boosted decision trees (GBDT).

For each combination of task and model class, we compared four configurations: a baseline configuration, the baseline configuration supplemented with 330-dimensional PDFM embeddings, the baseline configuration supplemented with 10 American Community Survey (ACS)~\citep{bib13} measures (reflecting age, sex, race, median household income, unemployment, and population density), and the baseline configuration supplemented with both ACS and PDFM covariates. Bayesian baselines used spatial random effects with a population-based expected-count offset, adding temporal random effects for nowcasting, while GBDT baselines relied on population data, supplemented by lagged outcome counts for nowcasting (see Methods).

\subsubsection{Interpolation}

For spatial interpolation---where models must estimate 2023 county-level CVD death counts for held-out counties without access to those counties' historical mortality records---Bayesian spatial models substantially outperformed GBDT models across all configurations (Table~\ref{tab3}). Relative to the no-covariate baseline in the Bayesian spatial model (MAE: 48.44), adding ACS covariates significantly reduced mean absolute error (MAE: 42.07; p<0.01), while adding PDFM embeddings yielded an intermediate reduction in error (MAE: 44.95), with overlapping cross-validation ranges (36.53--45.94 vs. 40.59--50.12). In the GBDT models, neither ACS (MAE: 74.02) nor PDFM (MAE: 79.58) significantly improved upon the no-covariate baseline (MAE: 74.64); while ACS achieved a lower county-pooled MAE than PDFM alone in GBDT (p<0.001), their 95\% bootstrap intervals across holdout replicates overlapped substantially (60.07--98.91 vs. 63.74--107.56). Because models predict Poisson-distributed counts of death across counties ranging from 500 to nearly 10 million residents, RMSE can be heavily influenced by prediction errors in a small number of large-population counties---as suggested by the magnitude of the RMSE/MAE ratios. None of the RMSE differences between covariate sets are statistically significant, and RMSE/MAE ratios are similar. Finally, unlike in longitudinal nowcasting (which trains on 10 years of county observations), combining all 330 PDFM dimensions with the 10 ACS covariates in a single cross-sectional year (Baseline + ACS + PDFM) did not improve interpolation accuracy over ACS alone (Bayesian MAE: 47.77 vs. 42.07; RMSE: 179.09 vs. 125.89), likely reflecting increased estimation variance when fitting 340 covariates to a single year of county data without feature reduction (Table~\ref{tab3}).

\subsubsection{Nowcasting}

For temporal nowcasting (Table~\ref{tab4}), models using single-month PDFM embeddings achieved predictive accuracy statistically indistinguishable in pairwise comparisons from models using 10 curated ACS socioeconomic measures---which require 60 months of pooled survey data and up to a 12-month release delay---establishing that off-the-shelf geospatial representations can match traditional census covariates while eliminating multi-year survey lags. Using ACS and PDFM in conjunction generally led to the strongest overall performance. When examining the covariates individually, both ACS and PDFM embeddings yielded similar point errors in Bayesian models relative to baseline (MAE: 24.30 and 24.28 vs. 23.99), whereas combining both ACS and PDFM covariates yielded a nominal improvement in both MAE (23.59) and RMSE (69.62 vs. 71.22). In GBDT models, both ACS (MAE: 19.13; RMSE: 57.69) and PDFM (MAE: 18.66; RMSE: 46.00) significantly outperformed the baseline model (MAE: 26.18; RMSE: 216.38; p<0.05), with PDFM producing the lowest MAE and RMSE among individual covariate configurations and the combined ACS + PDFM model achieving the lowest overall MAE (18.52; p<0.01; Table~\ref{tab4}).

Achieving nowcasting performance on par with ACS covariates is notable given the operational advantages of foundation model representations. Standard county-level ACS measures, inclusive of smaller population counties (<65,000 residents), rely on five-year pooling (aggregating 60 months of survey data) and are updated on an annual cadence, with processing and release up to 12 months after data collection~\citep{bib14}. In contrast, PDFM embeddings are computed from a single month of data and refreshed on a monthly cadence, providing immediate characterization of local conditions without the 60-month pooling and release delays of ACS.

\begin{table}[ht!]
\caption{Spatial interpolation results showing mean MAE and pooled mean RMSE (in county-level counts of cardiovascular disease deaths; lower values indicate better accuracy) from 20\% county holdout tests in 2023. Values in parentheses denote the minimum--maximum range across 5 cross-validation folds for Bayesian models and the 2.5th--97.5th percentile interval across 100 bootstrap replicates for GBDT models. Bold reflects the lowest MAE and RMSE across all models. ** indicates p<0.01 versus Baseline, and \textasciicircum\textasciicircum\textasciicircum{} indicates p<0.001 for Baseline + ACS versus Baseline + PDFM (clustered sign-flip test significant and clustered bootstrap 95\% confidence interval for paired error differences excludes zero); all other differences from Baseline and pairwise differences between ACS and PDFM configurations are not statistically significant.}\label{tab3}%
\fontsize{7.5}{9}\selectfont
\setlength{\tabcolsep}{0pt}
\begin{tabular*}{\textwidth}{@{\extracolsep\fill}>{ \raggedright\arraybackslash }p{0.18\textwidth}cccc@{}}
\toprule
 & \multicolumn{2}{c}{\textbf{Bayesian spatial model}} & \multicolumn{2}{c}{\textbf{Gradient boosted decision tree model}} \\
\cmidrule(lr){2-3}\cmidrule(lr){4-5}
\textbf{Model configuration} & \textbf{MAE} & \textbf{RMSE} & \textbf{MAE} & \textbf{RMSE} \\
\midrule
Baseline model & 48.44 (44.36, 52.87) & 126.87 (111.42, 153.35) & 74.64 (62.10, 94.37) & 368.57 (171.94, 565.82) \\
 \hspace{3mm} + ACS & \hspace{1mm} \textbf{42.07}$^{**}$ (36.53, 45.94) & \textbf{125.89} (87.04, 152.19) & \hspace{1mm} 74.02\textsuperscript{\textasciicircum\textasciicircum\textasciicircum} (60.07, 98.91) & 406.02 (174.89, 585.12) \\
\hspace{3mm} + PDFM & 44.95 (40.59, 50.12) & 133.70 (96.90, 165.19) & \hspace{1mm} 79.58 (63.74, 107.56) & 411.00 (185.25, 597.87) \\
\hspace{3mm} + ACS + PDFM & 47.77 (38.87, 52.47) & \hspace{1mm} 179.09 (101.72, 245.96) & \hspace{1mm} 75.81 (59.55, 103.32) & 403.91 (166.70, 583.31) \\
\botrule
\end{tabular*}
\end{table}

\begin{table}[ht!]
\caption{Nowcasting results showing MAE and pooled RMSE (in county-level counts of cardiovascular disease deaths; lower values indicate better accuracy) for the 100\% county holdout test in 2023 (trained on 2013--2022). Values in parentheses for GBDT models denote the 2.5th--97.5th percentile interval across 100 random-seed replicates. Bold reflects the lowest MAE and RMSE across all models. For GBDT models, * indicates p<0.05 and ** indicates p<0.01 versus Baseline (clustered sign-flip test significant and clustered bootstrap 95\% confidence interval for paired error differences excludes zero); pairwise differences between GBDT ACS and PDFM configurations are not statistically significant.}\label{tab4}%
\footnotesize
\setlength{\tabcolsep}{0pt}
\begin{tabular*}{\textwidth}{@{\extracolsep\fill}lcccc@{}}
\toprule
 & \multicolumn{2}{c}{\textbf{Bayesian spatial model}} & \multicolumn{2}{c}{\textbf{Gradient boosted decision tree model}} \\
\cmidrule(lr){2-3}\cmidrule(lr){4-5}
\textbf{Model configuration} & \textbf{MAE} & \textbf{RMSE} & \textbf{MAE} & \textbf{RMSE} \\
\midrule
Baseline model & 23.99 & 71.22 & 26.18 (25.51, 27.17) & \hspace{1mm} 216.38 (193.67, 240.93) \\
\hspace{3mm} + ACS & 24.30 & 72.83 & 19.13$^{*}$ (18.63, 19.86) & 57.69$^{*}$ (48.79, 71.58) \\
\hspace{3mm} + PDFM & 24.28 & 72.65 & 18.66$^{*}$ (18.43, 19.22) & \textbf{46.00}$^{*}$ (44.96, 47.61) \\
\hspace{3mm} + ACS + PDFM & 23.59 & 69.62 & \hspace{0.5mm} \textbf{18.52}$^{**}$ (18.39, 19.15) & 46.02$^{*}$ (45.14, 49.18) \\
\botrule
\end{tabular*}
\end{table}

These results carry direct implications for public health resource allocation. Cardiovascular disease kills approximately 916,000 Americans annually, yet final unsuppressed county-level mortality counts from the National Vital Statistics System may lag 1--2 years behind real time, and ACS covariates compound this delay, as they are drawn from multi-year rolling surveys that reflect socioeconomic conditions 2--3 years in the past (for counties with <65,000 residents). As both traditional census covariates and PDFM significantly improved the predictive power of machine learning nowcasting models---with no statistically significant difference between PDFM and ACS in head-to-head comparisons---the findings suggest that PDFM embeddings may offer a viable, timely alternative when up-to-date demographic survey data are unavailable, especially considering that they can be computed from same-year or same-month data without survey release delays. The results indicate that the strongest nowcasting performance can be achieved by leveraging a combination of the covariates, here using census data from 2019--2023 and PDFM from October 2023. Continued research should explore longitudinal performance with monthly and annual PDFM.

\subsection{Case Study 3: Population dynamics covariates improve forecasting in active vector-borne outbreak hotspots}

Effective management of seasonal infectious diseases like dengue requires forecasting models that can navigate complex environmental drivers and human dynamics across diverse geographic scales~\citep{bib15}. Timely, localized forecasts are essential for proactive vector control, yet previous studies in Mexico have shown that predictive value beyond short-term and seasonal temporal structure remains limited~\citep{bib16}. Moving from chronic and immunization surveillance to dynamic infectious disease outbreaks, this case study evaluates whether PDFM embeddings can enhance probabilistic dengue forecasting across thousands of municipalities---and specifically tests at which forecast horizons and local transmission regimes geospatial context provides the greatest operational benefit.

We produced probabilistic forecasts of new dengue cases across approximately 2,450 Mexican municipalities (GADM Administrative Level 2 units) at horizons of 1, 3, and 6 months over the period January 2020 to August 2025. A baseline TimesFM model~\citep{bib17} was compared with PDFM-adjusted TimesFM forecasts, in which a static embedding snapshot was incorporated through horizon- and quantile-specific residual models across the evaluation period. Performance was further stratified by municipality-level case burden to characterize where improvements arose.

\begin{figure}[htbp]
\centering
\includegraphics[width=\textwidth]{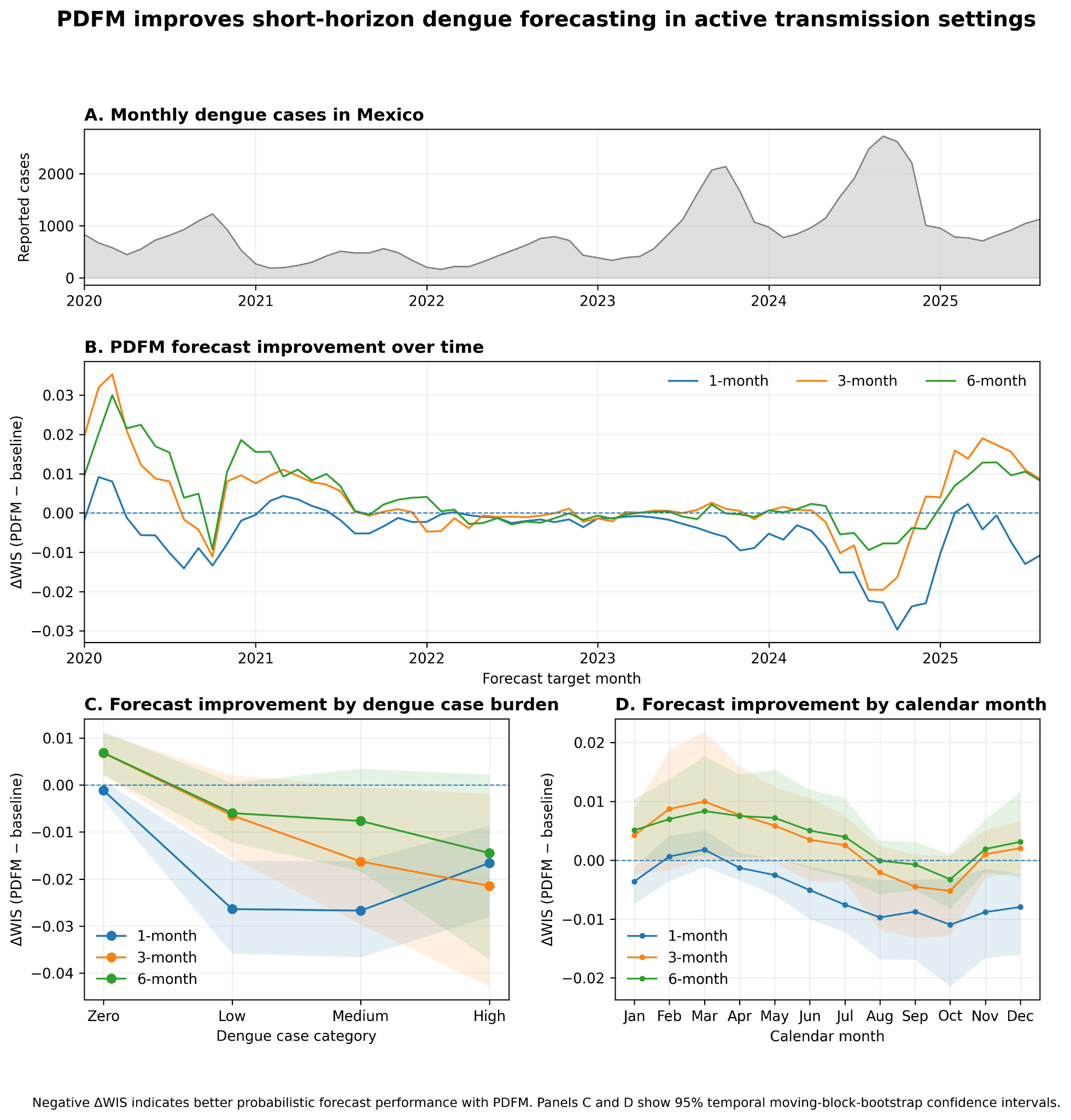}
\caption{PDFM improves short-horizon dengue forecasting during active transmission in Mexico. A) Monthly reported dengue cases during the January 2020--August 2025 evaluation period. B) Mean difference in Weighted Interval Score ($\Delta\text{WIS} = \text{WIS}_{\text{PDFM}} - \text{WIS}_{\text{baseline}}$) between PDFM-adjusted and baseline forecasts by forecast target month for 1-, 3-, and 6-month forecast horizons. Negative values indicate improved probabilistic forecast performance with PDFM. C) Mean $\Delta$WIS stratified by municipality-level dengue case burden. Municipality-months with zero reported cases are shown separately; non-zero observations are divided into low-, medium-, and high-case groups. Shaded regions indicate 95\% confidence intervals estimated using a temporal moving-block bootstrap with three-month blocks. D) Mean $\Delta$WIS by calendar month for 1-, 3-, and 6-month forecast horizons. Shaded regions indicate 95\% confidence intervals estimated using the same temporal moving-block bootstrap.}\label{fig4}
\end{figure}

Dengue activity varied substantially over the evaluation period, with particularly high national case burden during 2023--2024 (Figure~\ref{fig4}A). At the short horizon, PDFM significantly improved one-month forecast performance, with a mean $\Delta$WIS ($\text{WIS}_{\text{PDFM}} - \text{WIS}_{\text{baseline}}$) of $-$0.0051 (95\% temporal moving-block-bootstrap CI: $-$0.0081 to $-$0.0027; Diebold--Mariano p < 0.001), a result robust to bootstrap block lengths of 3, 6, and 12 months (Figure~\ref{fig4}B). At longer horizons, however, no robust improvement was observed: $\Delta$WIS was 0.0028 at three months (DM p = 0.140) and 0.0039 at six months (DM p = 0.074, sensitive to block length). The evidence therefore supports a clear short-horizon benefit that does not extend to longer forecast horizons.

To understand the reason for this short-horizon benefit, we retrospectively stratified forecasts according to the observed dengue case burden in each municipality for the target month. Municipality-months with zero observed cases in the target month were considered separately, while those with non-zero reported cases were divided into equal-frequency low-, medium-, and high-case groups (Figure~\ref{fig4}C). For one-month forecasts, PDFM added little value when no cases were present ($\Delta$WIS = $-$0.0012), but substantially improved predictions in all three active-transmission groups, with $\Delta$WIS ranging from $-$0.0172 to $-$0.0266 across the low-, medium-, and high-case strata. A similar pattern held at the three-month horizon despite the absence of an overall improvement: PDFM performed worse than baseline in zero-case municipality-months ($\Delta$WIS=0.0065) but improved forecasts in the medium- and high-case groups (Figure~\ref{fig4}C). These results indicate that the absence of an overall average improvement at longer horizons stems from two factors. First, at the three-month horizon, it masks substantial heterogeneity between municipalities with and without active transmission: because static PDFM embeddings capture a municipality's structural suitability for dengue, the residual model improves predictions when outbreaks occur in high-burden areas but introduces small over-predictions during quiet, zero-case months---which comprise the majority of municipality-months and dilute the overall average. Second, by the six-month horizon, forecasts span full seasonal transitions where incidence depends on future meteorological and vector-population shifts that have not yet occurred at the forecast origin and cannot be captured by static spatial covariates alone.

\begin{figure}[htbp]
\centering
\includegraphics[width=\textwidth]{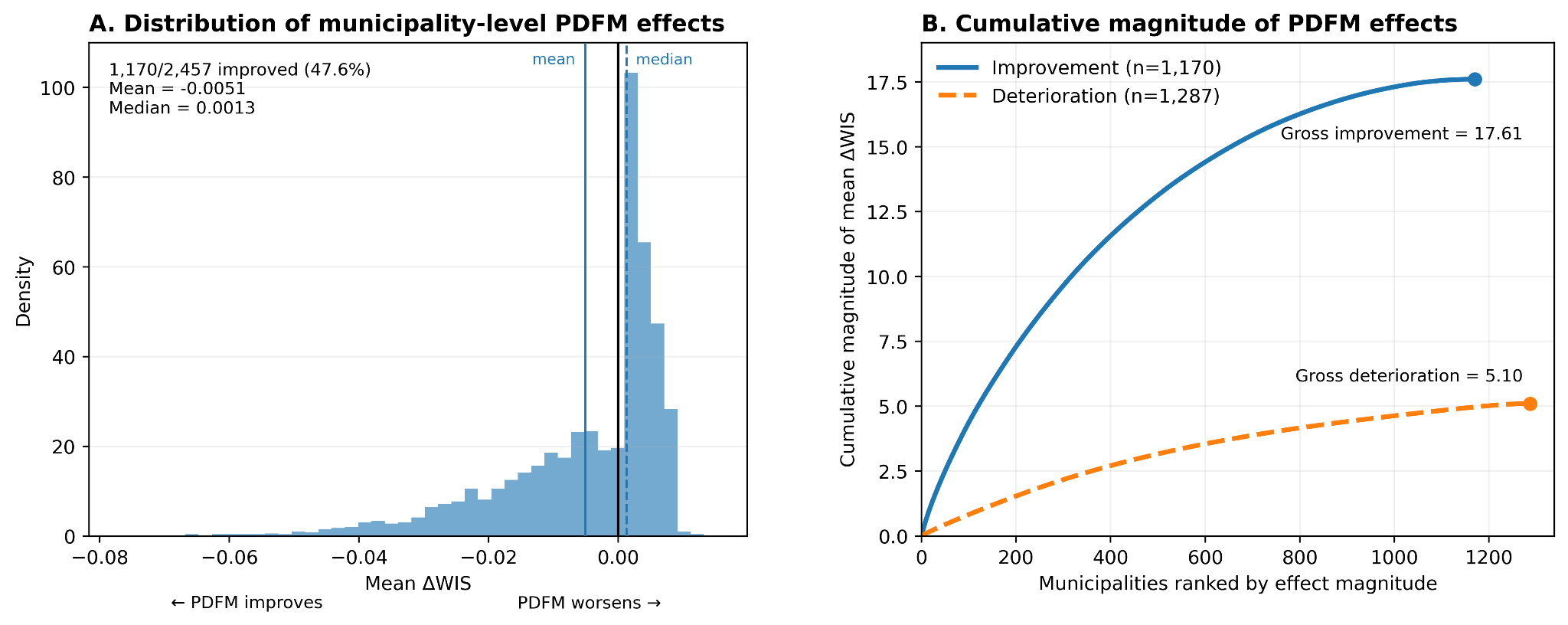}
\caption{Distribution and cumulative magnitude of time-averaged forecast accuracy differences across Mexican municipalities. A) Histogram displaying the difference in mean Weighted Interval Score ($\Delta\text{WIS}_m$) at Horizon 1 between the TimesFM model augmented with PDFM and the baseline TimesFM without PDFM. Differences are computed over the evaluation period from January 2020 to August 2025 across 2,457 municipalities. The solid black vertical line marks zero difference; negative values ($\Delta\text{WIS}_m < 0$) indicate a lower average forecast error for the model utilizing PDFM. The solid and dashed blue vertical lines indicate the mean and median, respectively. The distribution has a mean difference of $-$0.0051 and a median of 0.0013. Overall, the inclusion of PDFM improved average forecast performance in 47.6\% (1,170 out of 2,457) of the evaluated municipalities. B) Cumulative magnitude of municipality-level mean $\Delta$WIS after separately ranking municipalities with improved and worsened performance from largest to smallest effect magnitude. Gross improvement, defined as the sum of absolute mean $\Delta$WIS among municipalities with $\Delta$WIS < 0, was 17.61, compared with gross deterioration of 5.10 among municipalities with $\Delta$WIS > 0. Thus, although deterioration occurred in slightly more municipalities, improvements were substantially larger in magnitude.}\label{fig5}
\end{figure}

\begin{figure}[htbp]
\centering
\includegraphics[width=\textwidth]{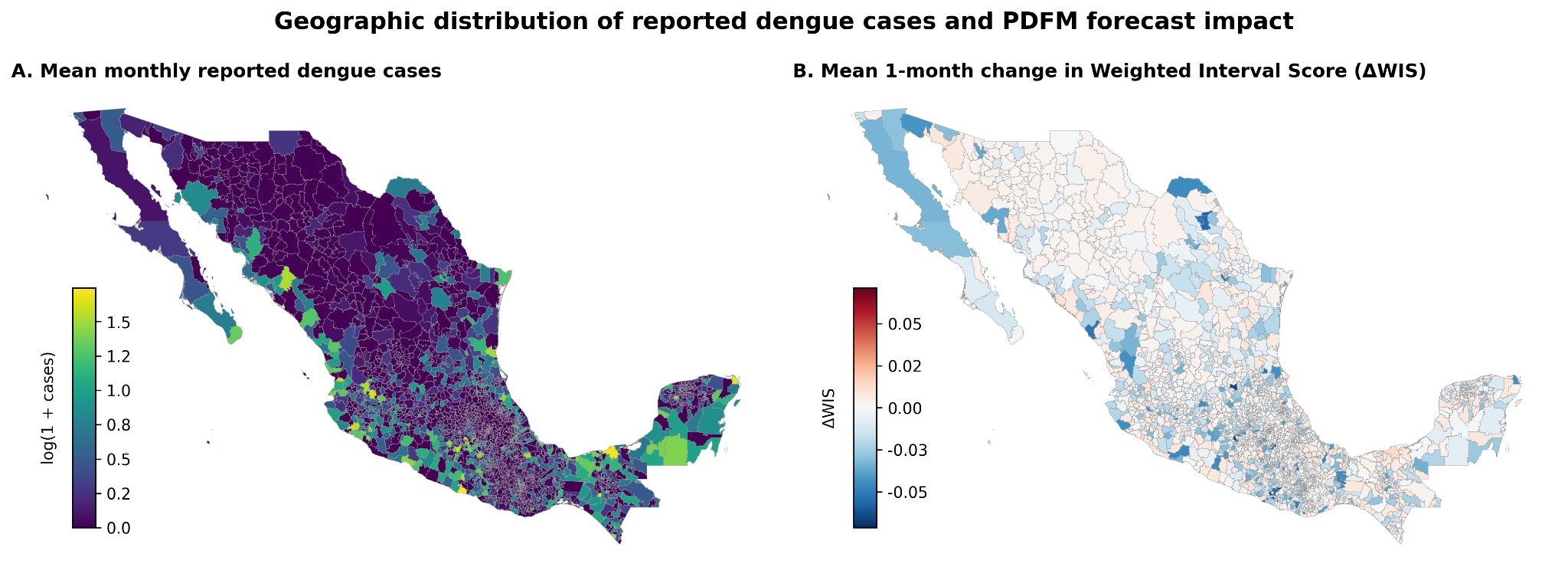}
\caption{Geographic distribution of reported dengue burden and change in Weighted Interval Score (WIS) applying PDFM embeddings in Mexico. A) Mean monthly reported dengue case burden by municipality over the evaluation period. Values are displayed as log(1 + mean monthly cases). B) Mean difference in Weighted Interval Score ($\Delta\text{WIS} = \text{WIS}_{\text{PDFM}} - \text{WIS}_{\text{baseline}}$) for one-month-ahead forecasts by municipality over the same evaluation period. Negative $\Delta$WIS indicates improved forecast performance with PDFM relative to baseline. Across 2,457 municipalities, 1,170 (47.6\%) had negative mean $\Delta$WIS; the median municipality-level $\Delta$WIS was 0.0013 (interquartile range $-$0.0111 to 0.0035), indicating that the overall improvement was driven by larger gains in a subset of municipalities rather than uniform improvement across locations.}\label{fig6}
\end{figure}

Forecast improvement also varied across the calendar year (Figure~\ref{fig4}D), consistent with temporal heterogeneity in dengue activity, though case burden provides a more direct characterization: the clearest improvements occurred when dengue was actively circulating within a municipality. The one-month improvement was also not uniformly distributed across municipalities---of 2,457 municipalities, 1,170 (47.6\%) showed improved average performance with PDFM, i.e., negative mean $\Delta$WIS. When stratified by retrospective case burden, the proportion of municipalities benefiting was substantially higher among municipalities with active transmission: 38.9\% of municipalities benefited in the zero-case stratum, compared with 71.9\%, 72.1\%, and 64.6\% in the low-, medium-, and high-case strata, respectively. Thus, higher dengue burden was associated with a greater proportion of municipalities benefiting from PDFM, although substantial heterogeneity remained even among high-burden municipalities. The median municipality-level $\Delta$WIS was slightly worse (0.0013) despite the overall improvement (Figure~\ref{fig5}A), indicating that larger gains among improving municipalities outweigh generally smaller deteriorations elsewhere (Figure~\ref{fig5}B). The geographic distribution of these improvements is heterogeneous (Figure~\ref{fig6}B), consistent with fine-scale spatial variation in Aedes-borne disease burden across Mexican municipalities~\citep{bib18} (Figure~\ref{fig6}A).

Ideally, PDFM embeddings would be temporally aligned with each forecast training period. However, given the lack of historical PDFM embeddings due to historical data retention restrictions, the primary analysis used the static October 2023 snapshot throughout. To assess sensitivity to the temporal snapshot used to construct the embeddings, we repeated the analysis using PDFM embeddings from October 2025 rather than October 2023, holding all other procedures fixed. The results were overall consistent: mean one-month $\Delta$WIS was $-$0.00505 (2025) versus $-$0.00514 (2023), and the case-burden stratification pattern was preserved. This observed robustness of the short-horizon benefit indicates that the embeddings capture durable geographic and environmental characteristics rather than transient conditions tied to a specific snapshot, supporting their use as reliable covariates across time.

Taken together, these analyses reveal a consistent pattern: rather than providing a uniform improvement across all municipalities and forecast horizons, PDFM delivers its clearest contribution for short-horizon forecasts in municipalities with active dengue transmission---a finding that is robust across embedding snapshots and geographic strata. This pattern is particularly valuable for operational dengue response, where timely identification of changing local risk informs vector-control and preparedness decisions~\citep{bib15}.

\subsection{Case Study 4: Population dynamics for hotspot prediction in data-scarce contexts}

Containing epidemic-prone enteric diseases such as cholera (\textit{Vibrio cholerae}) requires warnings weeks before surges overwhelm treatment centers. In resource-limited and conflict-affected regions, reporting delays, spatial friction, and missing infrastructure make this difficult. Complementing the monthly dengue evaluation in Case Study 3---where PDFM improved forecasts at the 1-month horizon before seasonal weather transitions dominated at 3 to 6 months---this case study examines weekly lead times from 1 to 8 weeks in a low-resource setting to pinpoint when geospatial context begins to add value over recent case history. Specifically, we test whether adding a custom, low-resource 200-dimensional version of PDFM embeddings to historical surveillance features improves prospective cholera emergence forecasts across 403 health zones in the DRC.

We trained models on national surveillance data from before 1 November 2023 to prevent data leakage from the PDFM embeddings encoding data from October 2023. We then tested them prospectively over the next 89 weeks (through 14 July 2025), at lead times of 1, 2, 4, and 8 weeks. Drawing from prior work, an emergence event was defined as a three-week window with $\ge$100 suspected cases, starting from a quiet week (<50 cases)~\citep{bib24}. Such events were rare, making up 0.92\% of zone-weeks at a 1-week lead (323 events) and 2.40\% at an 8-week lead (776 events), simultaneously illustrating both the difficulty and importance of the task.

Consistent with the 1-month benefit observed for dengue (Case Study 3), the utility of the embeddings followed a clear temporal window across forecast horizons (Table~\ref{tab5}). At immediate 1--2 week horizons, autoregressive recent case history was already sufficient and both feature sets performed similarly; however, at 4- and 8-week horizons---the exact 1-to-2-month operational window after immediate case persistence decays, when response teams must pre-position supplies---adding PDFM embeddings yielded significant improvements, with PR-AUC rising from 0.2832 to 0.3107 (4 weeks) and from 0.2733 to 0.2899 (8 weeks).

In practice, response teams do not work from calibrated probabilities. They work from a short list of high-risk zones, which they use to pre-position treatment kits and oral rehydration points. We therefore also measured Precision@5: the share of each week's five predicted top-ranked zones that went on to actually have an emergence event (even rows in Table~\ref{tab5}). On this measure, the embeddings helped more, and the gains were statistically significant at both longer horizons. At 4 weeks, Precision@5 rose from 0.3435 to 0.3906. At 8 weeks, it rose from 0.3556 to 0.4198. This worked out to 2.10 true events caught per week among the top five zones, compared with 1.78 using case history alone. At one week, by contrast, Precision@5 was slightly lower with embeddings ($\Delta$ = $-$0.0227, $-$8.0\%), though this drop was not significant under the permutation test (p = 0.103).

\begin{table}[htbp]
\caption{Prospective emergence forecasting performance, all health zones. History+PDFM versus History-only, predictions pooled over five random seeds. Lead time is the prediction horizon. $\Delta$ is History+PDFM minus History-only. CIs are unadjusted percentile bootstraps while p-values are Benjamini--Hochberg adjusted within this table.}\label{tab5}%
\footnotesize
\setlength{\tabcolsep}{0pt}
\begin{tabular*}{\textwidth}{@{\extracolsep\fill}llccccc@{}}
\toprule
\textbf{Lead Time} & \textbf{Metric} & \textbf{History-only} & \textbf{History+PDFM} & \textbf{$\Delta$ (\%)} & \textbf{95\% CI} & \textbf{\textit{p}} \\
\midrule
1 wk & PR-AUC & 0.2931 & 0.2969 & $+$0.0038 ($+$1.3) & $-$0.0161 to $+$0.0258 & 0.72 \\
 & Precision@5 & 0.2841 & 0.2614 & $-$0.0227 ($-$8.0) & $-$0.0432 to $-$0.0023 & 0.103 \\
2 wk & PR-AUC & 0.2990 & 0.2860 & $-$0.0129 ($-$4.3) & $-$0.0304 to $+$0.0055 & 0.23 \\
 & Precision@5 & 0.3011 & 0.3126 & $+$0.0115 ($+$3.8) & $-$0.0138 to $+$0.0368 & 0.52 \\
4 wk & PR-AUC & 0.2832 & 0.3107 & $+$0.0275 ($+$9.7) & $+$0.0089 to $+$0.0479 & 0.019 \\
 & Precision@5 & 0.3435 & 0.3906 & $+$0.0471 ($+$13.7) & $+$0.0141 to $+$0.0776 & 0.019 \\
8 wk & PR-AUC & 0.2733 & 0.2899 & $+$0.0166 ($+$6.1) & $+$0.0011 to $+$0.0321 & 0.064 \\
 & Precision@5 & 0.3556 & 0.4198 & $+$0.0642 ($+$18.1) & $+$0.0296 to $+$0.1037 & 0.013 \\
\botrule
\end{tabular*}
\end{table}

We further stratified health zones by historical endemicity based only on data from before the November 1, 2023 evaluation cutoff. This resulted in 15 endemic, 340 sporadic, and 48 dormant zones (Extended Data Table~\ref{exttab2}). Our results show the largest gains in those endemic zones. Specifically, PR-AUC at 4 weeks rose significantly from 0.4209 to 0.4729, while at 8 weeks, the Precision@5 exhibited the biggest gains, rising from 0.3333 to 0.3975, with a better shortlist in 23 weeks and a worse one in only 3. Sporadic zones showed a similar 4-week PR-AUC gain, from 0.2127 to 0.2397. However, the gain in sporadic zones fell just short of significance after correcting for multiple comparisons across strata (p = 0.059).

\subsection{Case Study 5: Scope conditions of geographic context in individual-level clinical risk prediction for postpartum depression}

Postpartum depression (PPD) is common and rising. In the United States, 11.9\% of mothers reported depressive symptoms in 2023~\citep{bib19}, compared with 12.5\% in the 2012--2021 cohort (survey-weighted) used in this analysis. Diagnosed PPD doubled from 9.4\% to 19.0\% between 2010 and 2021 in a large US health system~\citep{bib20}. Worldwide, global meta-analyses estimate PPD at 17--19\%~\citep{bib21,bib22}, with regional estimates from 11\% (Oceania) to 40\% (Southern Africa). Many affected mothers are not identified or treated. To establish the scope conditions and boundaries of geospatial foundation models, we next move from population-level surveillance to individual-level clinical risk stratification. We tested three boundary questions: whether coarse place-level PDFM embeddings add a transferable signal to individual-level PPD risk prediction in unseen states, whether they can substitute for patient-level socioeconomic records (household income and insurance), and whether geography alone can narrow within-place performance gaps across race, ethnicity, and urbanicity.

We analyzed 332,970 respondents to the CDC Pregnancy Risk Assessment Monitoring System (PRAMS; births 2012--2021). Each mother was linked to the population-weighted PDFM embedding of her state $\times$ urban/rural location, the finest geography the research file allows, yielding 87 locations.

The analysis plan was pre-registered before any model saw outcomes, and the test sets were sealed in advance. This analysis supersedes an earlier analysis of these data~\citep{bib23}. Models were evaluated in three settings: mothers in states seen in training (S1, primary); states never seen in training (S2, ten jurisdiction-grouped folds); and later years (S3, 2020--21). Comparators comprised a baseline of individual-level predictors, the baseline augmented with individual household income and Medicaid status, and random- and shuffled-embedding controls. Race and ethnicity defined evaluation groups but were never used as model inputs.

The prediction gain was small but measurable. In the primary setting, adding PDFM improved discrimination ($\Delta$AUC +0.0020, 0.0008--0.0031; Holm-adjusted p = 0.012), on a baseline AUC of 0.62. In states never seen in training, where the model has no other geographic information, the gain was larger (+0.0038, 0.0004--0.0072). Real embeddings outperformed random embeddings (+0.0047, 0.0013--0.0081), which shows the transferred signal comes from PDFM's content. The gain did not persist forward in time when models trained on 2012--2019 births were applied to 2020--2021 births (S3: +0.0006, $-$0.0005 to 0.0015).

PDFM embeddings did not substitute for individual socioeconomic data. A model with individual income and Medicaid status discriminated better than one with PDFM alone ($\Delta$AUC $-$0.0116, $-$0.0144 to $-$0.0089). PDFM recovered 15\% (6--24\%) of the gain from individual socioeconomic data, and added little beyond it (+0.0011, 0.0000--0.0022; Holm-adjusted p = 0.10).

None of the 21 pre-registered race $\times$ ethnicity $\times$ urbanicity performance gaps closed in seen or unseen states. Incorporating PDFM embeddings mainly changed how rural mothers were screened, with the direction depending on the screening policy. Under a fixed screening capacity, in which the 20\% of mothers at highest predicted risk are flagged for follow-up, the PDFM model found an estimated 5,640 more rural PPD cases per year (2,576--8,859), at the cost of 23,474 more false alarms (8,964--39,309). The shift toward rural mothers held at every capacity tested (10\%, 20\% and 30\% flagged). Under a fixed sensitivity target of 80\%, by contrast, PDFM averted 17,723 false alarms (6,952--32,033) while flagging 1,722 fewer rural cases (650--3,127).

\begin{figure}[htbp]
\centering
\includegraphics[width=\textwidth]{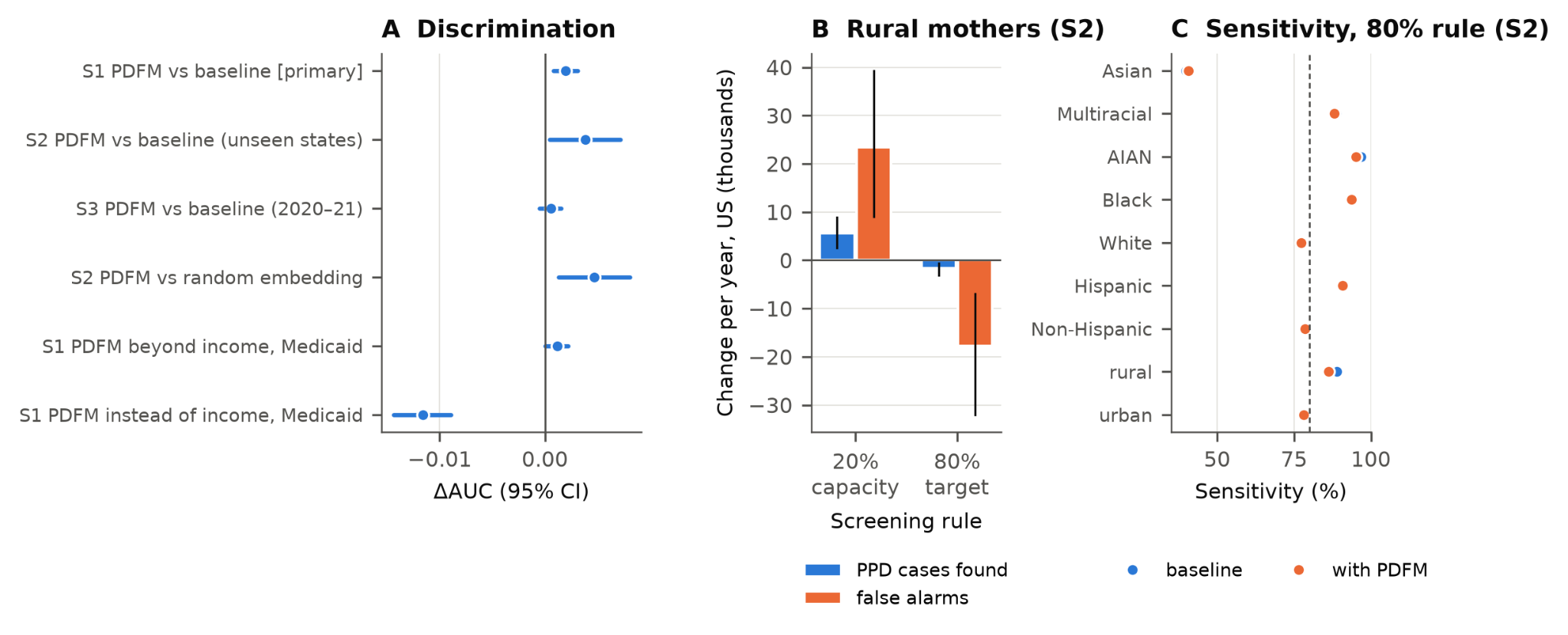}
\caption{PDFM in individual-level postpartum-depression risk prediction (PRAMS, 332,970 mothers). A) Change in discrimination ($\Delta$AUC, 95\% CI) from adding PDFM embeddings. Rows: seen states (S1, pre-registered primary); unseen states (S2); later years (S3); real vs random embedding in unseen states; PDFM added beyond individual income and Medicaid; and PDFM in place of them. B) Rural mothers in unseen states (exploratory; US-weighted, per year): change in PPD cases found and in false alarms under two screening policies. Fixed screening capacity flags the 20\% of mothers at highest predicted risk. The fixed target uses a threshold set on training data for $\ge$ 80\% sensitivity. C) Sensitivity by group under the 80\% rule, without (blue) and with (orange) PDFM. The dashed line marks 80\%. Error bars are 95\% bootstrap CIs; S2 CIs resample jurisdictions.}\label{fig7}
\end{figure}

By contrast, the largest baseline demographic disparity---the under-detection of Asian mothers under the 80\% sensitivity target, where all models detected fewer than half of cases (39.8\%)---was left unchanged with or without PDFM (Figure~\ref{fig7}C).

Taken together, these results indicate that at state $\times$ urban/rural resolution, geographic context redistributes predicted risk between places, whereas most of the predictive signal, and the largest disparities, reside within places.

To gauge what these effects would imply beyond the United States, we conducted exploratory transport scenarios using a screening simulator that reconstructs a target population with a specified rural share and PPD prevalence and re-applies the screening rule within it (Methods; Extended Data Table~\ref{exttab5}). Prevalence across scenarios ranged from 13.5\% to 40\%~\citep{bib21}, and the simulator reproduces the US estimates exactly when run at US composition. Under a fixed 20\% screening capacity, with the same direction observed at 10\% and 30\% in every scenario, PDFM shifted detection toward rural mothers, and the magnitude of the shift increased with rural share and prevalence: +337 rural cases per 100,000 births worldwide (137--545), +454 in Southern Asia (65\% rural, 22\% prevalence; 138--692) and +716 in Southern Africa (36\% rural, 40\% prevalence; 312--1,118). Because fewer urban cases were detected in exchange, the net change in total cases found was not distinguishable from zero in any scenario except South America, where the lower confidence bound only marginally excluded zero (+157 per 100,000 births; +1 to +305). Scaled to 132 million annual births, this corresponds to approximately 445,000 additional rural cases detected per year (181,000--720,000) and 283,000 fewer urban cases (86,000--500,000). Under the fixed 80\% sensitivity target, the predominant effect was the avoidance of false alarms, approximately 1.5 million fewer per year worldwide (0.01--3.2 million), with no distinguishable change in total cases found. These scenarios assume that US score distributions transfer to other settings; because both PDFM v0 and PRAMS are restricted to the United States, this assumption cannot be tested here.

This analysis features a number of limitations. The first of these is the time gap between the embeddings and the PRAMS births: while the former encodes data from 2022 to 2024, the latter features data from 2012 to 2021. The gain did not vary with the length of this gap (slope 0.0000/yr, $-$0.0004 to 0.0004). The earliest available version of the PDFM embeddings encodes data from October 2023, which also postdates all births. A comparison with it is left to follow-up work. In addition, Alaska and Hawaii are absent from the PDFM v0 release that was used, which excludes most Native Hawaiian and Pacific Islander respondents. The pre-registered label-permutation control fell just outside its expected range (AUC 0.475). A post-hoc null of 20 further permutations centered on 0.499 (95\% range 0.479--0.522), consistent with chance.

\section{Discussion}

This work illustrates that planetary geospatial foundation models represent a viable new paradigm for public health surveillance, one in which pre-trained representations of place can support diverse disease prediction tasks without requiring task-specific training. Using PDFM as a proof-of-concept, we demonstrate this potential through five independent case studies across four countries. Our findings show that geospatial covariates encoding search trends, mobility patterns, and environmental data can bridge critical gaps in the public health data landscape, where information is often sparse, infrequently updated, or confined to national boundaries.

The practical utility of this paradigm is best understood not through the aggregate statistical improvement but rather through the public health decisions it informs. Across the cases examined here, the greater value lies in how even modest error reductions translate into materially better allocation of scarce public health resources, whether by improving resource targeting or correcting high-stakes outlier predictions. In the US--Canada border analysis, domestic-only models failed to capture cross-border behavioral patterns influencing immunization rates, and incorporating Canadian contextual covariates yielded a 36\% relative increase in explained variance for near-border vaccination prediction. For the approximately 18.6 million people in these 146 border counties, this gain is particularly significant given that national MMR coverage has declined to 92.7\% amid the largest US measles resurgence since 1992, underscoring that epidemiologic environments in border regions are shaped by spatial dependence extending beyond national boundaries. For cardiovascular disease nowcasting, models augmented with PDFM embeddings and those augmented with American Community Survey (ACS) data showed no statistically significant differences in their respective outputs, lending credence to PDFM's role as a more timely and easily available alternative.

Beyond improving average accuracy, PDFM covariates disproportionately corrected predictions where errors are most costly. In the dengue analysis in Mexico, rather than uniformly improving predictions across all locations, PDFM covariates disproportionately corrected large forecasting errors in high-burden outbreak hotspots---the precise locations where accurate short-term predictions are most valuable for outbreak response. In cholera hotspot forecasting in the DRC, supplementing historical data with PDFM embeddings led to boosts in performance especially in endemic health zones.

Across 332,970 US mothers, the largest effect of adding geographic context was a policy-dependent reallocation of screening among rural mothers, a reminder that population-level context helps most where individual data are missing. Scale changes what a small individual-level gain means. In a verified screening simulator that rebuilds populations with the rural shares and postpartum-depression prevalence of other world regions, the same US-trained, location-aware model does two different things depending on the screening policy. With fixed screening capacity, it moves detection toward rural mothers: several hundred more rural cases per 100,000 births in rural-majority, high-prevalence settings, at the cost of fewer urban cases. With a fixed detection target, it mainly avoids false alarms, which carry their own burden for families and health systems. Which of these is a benefit is a policy choice, not a model property. Rather than letting a place-aware model shift screening implicitly, programs could allocate capacity to rural and urban mothers explicitly and use place context for population-level outreach, reserving individual detection for instruments that carry individual-level information, such as validated questionnaires or continuous signals from wearable devices. Testing it outside the US will need embeddings and outcome data from those settings. Following the idea of verified code world models~\citep{bib25}, we implemented these scenarios as an executable screening simulator that reproduces the observed US results exactly; such simulations make transport assumptions explicit and checkable.

These results represent a fundamental departure from the prevailing paradigm in disease modeling. Traditional approaches require substantial computational resources, extensive manual feature engineering, and large task-specific ground-truth datasets. These resources are rarely accessible during public health emergencies or in resource-limited settings. Geospatial foundation models instead generate generalizable, low-dimensional location representations through self-supervised pre-training on multimodal, privacy-preserving data streams. These versatile covariates can be integrated into existing analytical workflows across vastly different public health tasks, from vaccination uptake to chronic cardiovascular mortality to vector-borne and waterborne disease outbreaks, without requiring model retraining or specialized deep-learning expertise. This plug-in architecture is essential: geospatial foundation models should enhance existing tools, not demand that practitioners abandon proven methods.

Despite this versatility, geospatial foundation models are not a universal solution; their value depends critically on the temporal dynamics, spatial scale, and structural nature of the prediction task. For slowly changing, structurally driven health outcomes, such as interpolating county-level CVD mortality from cross-sectional data, traditional demographic surveys achieved nominally lower error (Bayesian spatial models with ACS achieved an MAE of 42.07 versus 44.95 for PDFM, though the difference was not statistically significant). The covariates show particular promise when rapid behavioral and environmental shifts drive disease outcomes, as in nowcasting and short-term forecasting, where they may capture dynamics that census data, with its multi-year update cycles, cannot. More broadly, geospatial foundation models complement rather than replace existing data systems. Their development and evaluation remain dependent on the quality of underlying outcome data, and improvements in the completeness, timeliness, and spatiotemporal resolution of these data could further strengthen their utility.

The primary technical limitation of the current evaluations is the use of static rather than time-varying geospatial embeddings. Although standard PDFM embeddings are generated from a single month of data and refreshed on a monthly cadence, historical data retention constraints restricted our retrospective analyses to single temporal snapshots (primarily October 2023) rather than longitudinal embedding sequences aligned with each training and forecast window. In the dengue analysis, static embeddings captured durable local susceptibility---improving one-month forecasts and three-month forecasts in active-transmission municipalities---but slightly over-predicted in zero-case months at three months (diluting the overall average) and could not anticipate seasonal transitions at six months. Two observations from this work suggest the potential value proposition of longitudinally available embeddings: the dengue benefit was unchanged when the October 2023 snapshot was replaced with one from October 2025, and in the postpartum-depression analysis the gain did not vary with the interval between embedding and birth. These indicate that the cross-sectional structure the embeddings encode is stable over at least two years. This is reassuring for the retrospective designs used here, and it also means that the operational advantage of monthly refresh is a hypothesis about within-location change during periods of rapid shift, rather than a result these experiments demonstrate. Notably, this boundary provides direct motivation for an important future direction: evaluating longitudinal sequences of monthly refreshed geospatial covariates matched directly to disease outcome periods to capture dynamic behavioral and environmental shifts alongside baseline structural risk. However, reliance on digital signals such as search trends and mobility data also introduces a trade-off between data sparsity and temporal resolution in lower-connectivity settings. In the DRC cholera analysis, for example---where internet penetration was 19.7\% of the population in 2024~\citep{bib26}---sparse search activity required pooling search data over a six-month window and compressing features via PCA rather than using a single-month snapshot, a constraint that limits how rapidly longitudinal embeddings can update to reflect acute shocks or humanitarian crises. Incorporating alternative, higher-penetration data modalities and geographic transfer learning may help improve both coverage and temporal responsiveness in data-scarce populations.

The use of static embeddings also presents a seeming contradiction in PDFM's benefit for short-term forecasts when predicting dengue and its long-term forecasts when predicting cholera hotspots. However, this contradiction is easily reconciled by translating the disease-relative time horizons to absolute time. Dengue forecasts showed statistically significant improvement at one month but not beyond that time horizon. Cholera forecasts improved at four and eight weeks but not at one or two. In both settings the embeddings therefore added information in a window of roughly one to two months. We hypothesize that this window is long enough that the persistence of recent case counts had decayed, yet short enough that the seasonal and meteorological transitions governing longer horizons had not yet taken hold. The recurrence of this window across a monthly count target and a weekly binary emergence target suggests a general property of the embeddings rather than a disease-specific effect.

Another direction for future work is to compare how interpretable PDFM embeddings are relative to traditional covariates. Each traditional census covariate represents a specific demographic or socioeconomic construct, such as median household income or the percentage of unemployed residents. In contrast, an individual PDFM embedding dimension has no clear meaning of its own and can only be traced back to the input modality it came from (aggregated search trends, the built environment, or environmental factors). This makes the embeddings less suited to descriptive or causal analyses that try to identify which place-based factors drive differences in health outcomes. Such analyses are a central use of labeled covariates like ACS measures in epidemiological research. Future work could address this limitation in two ways. One is to measure how much of the variation in labeled census measures the embeddings capture. The other is to attribute predictive gains to the blocks of dimensions that belong to each modality.

Finally, all applications except Case Study 5 operate at the population level. A natural extension is dyadic modeling that fuses PDFM's place-based context with individual-level physiological or behavioral signals, for example from wearable devices. Population-level place context and personal sensing carry largely orthogonal information; Case Study 5 shows the limit of place alone. Complementarity should not be assumed. In a perinatal screening study~\citep{bib27}, geospatial context that is informative at the population level carried no signal when used to rank individuals, whereas a speech model applied to the same women did. Fusion pays off only where each component supplies information the other lacks, and that has to be tested empirically. We are now doing this directly in pregnancy: a prospective study will pair continuous wearable measures, such as sleep, activity and heart-rate patterns, with PDFM's place context to ask whether personal and population-level signals are complementary for tracking perinatal mental health.

These findings collectively establish that planetary geospatial foundation models can bridge data gaps, capture complex human--environment interactions across sovereign borders, and provide a critical safety net during system shocks when traditional surveillance fails, facilitating a shift from reactive, task-specific approaches to proactive global health surveillance. Critically, the improvements documented here are not merely statistical; if they are utilized properly, they could affect which counties are flagged for vaccination outreach during measles outbreaks, how timely the data used to predict cardiovascular mortality is, whether vector control teams are deployed before or after dengue case surges breach endemic channel alert thresholds, which health zones are targeted for proactive resource deployment during cholera outbreaks, and how much the rural/urban divide affects postpartum depression screening. The capacity to deploy pre-trained, generalizable geospatial covariates into existing epidemiological workflows without task-specific retraining addresses a fundamental scalability challenge in global health, offering a complementary layer of intelligence for both routine surveillance and urgent outbreak response.

\section{Methods}

\subsection{PDFM Embeddings}

\subsubsection{Data Sources}

PDFM captures population dynamics by integrating three primary data modalities at fixed geographic administrative levels for each country. Originally detailed in~\citet{bib4}, the model incorporates the following datasets:

\begin{itemize}
\item Aggregated Search Trends: Privacy-preserving counts of Knowledge Graph entities~\citep{bib28} appearing in web searches (e.g., specific symptoms or clinical resources). The use of Knowledge Graph builds on the original Search Trends signal by mapping the features to broader categories, providing higher privacy and aggregation guarantees and improving the signal in data-strapped regions of the world~\citep{bib4}.
\item Built Environment: Geospatial density of 1,192 Point of Interest (POI) categories (e.g., clinical facilities, pharmacies) and aggregated ``place visit counts,'' or how busy the POI categories are based on aggregated mobility patterns for 683 POI categories.
\item Environmental Determinants: Meteorological data and high-resolution air quality measurements (including PM\textsubscript{2.5}, NO\textsubscript{2}, CO, ozone) provide ecological context.
\end{itemize}

The embeddings used for most analyses in this paper were computed using data from October 2023. The postpartum depression screening analysis used embeddings from the original PDFM publication (v0 public release; input data July 2022--mid-2024)~\citep{bib4}. To compensate for temporal misalignment in the dengue case study, a version of the embeddings computed using data from October 2025 was used in a sensitivity analysis. In the cholera case study, search data was aggregated over a 6-month period ending in October 2023 (inclusive).

\subsubsection{PDFM Architecture and Training}

The PDFM architecture is a deep neural network encoder where each geographic location is processed independently. For each location, the high-dimensional input features across all three modalities are concatenated and compressed to produce a 330-dimensional location embedding.

The model is pre-trained using a self-supervised learning (SSL) framework. Rather than being restricted to a single target task or relying on external supervision, the network learns the underlying baseline characteristics of a location by reconstructing its own input feature modalities from the shared bottleneck embedding. To encourage representational disentanglement and effectively allocate model capacity, the 330-dimensional embedding is partitioned using dataset-specific prediction heads:

\begin{itemize}
\item Dimensions 0--127 are allocated to reconstruct the aggregated search query distribution.
\item Dimensions 128--255 are shared by two distinct heads to reconstruct the POI category density and busyness histograms.
\item Dimensions 256--329 are allocated to reconstruct the environmental determinants.
\end{itemize}

Each task-specific head reconstructs its original input feature space. The network is trained end-to-end by optimizing a balanced, weighted sum of reconstruction losses across all heads simultaneously. This self-supervised objective ensures that the resulting embeddings are versatile, scalable, and generalizable mathematical representations of ``place.'' The resulting embeddings are utilized downstream as covariates for statistical modeling.

For the embeddings used in the cholera analysis, Principal Component Analysis was used to compress the input data into 200-dimensional embeddings. This was due to the DRC training data being an order of magnitude smaller than for other countries, making the above approach infeasible.

\subsection{Extrapolating MMR Vaccination Uptake on the US-Canada Border}

\subsubsection{Data}

The outcome variable is county-level measles--mumps--rubella (MMR) vaccination uptake among children under five years of age in the United States. County estimates were derived from a previously developed small-area estimation framework using survey responses collected between 2023 and 2024 through the OutbreaksNearMe (ONM) digital health surveillance platform, previously validated by the CDC. In brief, parent-reported vaccination status of children under five was modeled using a multilevel regression with poststratification (MRP) approach that combines individual-level survey responses with demographic population data to generate population-representative estimates for all U.S. counties~\citep{bib29,bib30}. These model-based county estimates represent the predicted proportion of children (under the age of five) receiving at least one dose of MMR vaccine and serve as the target variable for the predictive analyses conducted in this study.

\subsubsection{Experimental Setup}

To incorporate cross-border contextual information, we constructed distance-weighted contextual embeddings for each U.S. county (excluding Connecticut, as it consists of planning regions instead of counties) using Canadian Forward Sortation Areas (FSAs) located within a specified geographic radius of the county centroid. Embeddings from nearby Canadian FSAs were aggregated using inverse-distance weights, producing a contextual feature vector summarizing the surrounding Canadian environment. These aggregated cross-border embeddings were appended to the U.S. county embedding features and used as predictors in the cross-border prediction models. This formulation allows the model to incorporate contextual information from geographically proximate environments across the national boundary and evaluate whether such cross-border signals improve prediction of vaccination uptake in U.S. counties. The primary specification uses a 150 km radius, selected to capture plausible cross-border interaction zones while limiting the influence of distant regions.

For each U.S. county $i$, we construct a Canadian contextual embedding vector by aggregating nearby FSAs.

Let $d_{ij}$ denote the geographic distance between U.S. county $i$ and Canadian FSA $j$. Canadian FSAs within a radius $R$ are included:

\begin{equation}
d_{ij} \le R
\end{equation}

The Canadian embedding exposure for county $i$ is computed as an inverse-distance weighted average:

\begin{equation}
\tilde{C}_i = \sum_{j \in N(i)} w_{ij} C_j
\end{equation}

Where $N(i)$ denotes the set of Canadian FSAs within radius $R$ of county $i$.

The weights are defined as

\begin{equation}
w_{ij} = \frac{d_{ij}^{-1}}{\sum_{j \in N(i)} d_{ij}^{-1}}
\end{equation}

Distances are computed using projected coordinates (EPSG:3347), which preserve distance relationships across the U.S.-Canada border.

\subsubsection{Models}

We estimate ridge regression models predicting county vaccination rates. Ridge regression is used to stabilize estimation with high-dimensional correlated predictors.

The baseline model uses U.S. embeddings only:

\begin{equation}
y_i = \alpha + E_i \beta + \epsilon_i
\end{equation}

The cross-border model augments the feature space with Canadian contextual embeddings:

\begin{equation}
y_i = \alpha + E_i \beta + \tilde{C}_i \gamma + \epsilon_i
\end{equation}

Parameters are estimated using ridge regression, with $\theta = (\beta^{\top}, \gamma^{\top})^{\top}$ and $X_i = (E_i, \tilde{C}_i)$:

\begin{equation}
\min_{\alpha, \theta} \sum_i (y_i - \alpha - X_i \theta)^2 + \lambda \|\theta\|^2
\end{equation}

The regularization parameter $\lambda$ is selected via 10-fold cross-validation.

Three main model types are evaluated as follows:

\begin{itemize}
\item US-only model that predicts county vaccination uptake using only U.S. embeddings.
\item Canadian-only model that predicts U.S. county vaccination uptake using only Canadian contextual embeddings to test whether Canadian context alone contains predictive signal.
\item US and Canadian context model that includes both U.S. embeddings and distance-weighted Canadian contextual embeddings.
\end{itemize}

To identify which Canadian regions contribute most strongly to the cross-border predictive signal, we conducted a leave-one-out influence analysis. After estimating the cross-border predictive model, we evaluated the influence of each Canadian postal region by removing its weighted embedding from the contextual embedding of every U.S. county that included it and recomputing predictions while holding the fitted model coefficients fixed. Prediction error was defined as the absolute difference between the observed vaccination rate and the model's predicted value. We measured how much prediction error increased when a given Canadian region was removed relative to the full cross-border model, and defined the influence of that region as the total increase in prediction error across all affected counties. Figure~\ref{fig2} visualizes these results on both sides of the border: U.S. counties are shaded by the improvement in prediction accuracy when Canadian contextual embeddings are included, while Canadian postal regions are shaded by their influence in the leave-one-out analysis.

\subsubsection{Evaluation Metrics}

Model performance is evaluated using root mean squared error (RMSE), mean absolute error (MAE), Pearson correlation, and R$^2$.

Statistical significance was assessed using a permutation test in which Canadian contextual embeddings were randomly reassigned among border counties while preserving the modeling pipeline and cross-validation procedure. Observed improvements in predictive performance were compared with the resulting empirical null distribution. As a robustness analysis, matched 10-fold cross-validation was repeated 100 times using different random fold assignments. Improvements in predictive performance were reproduced across all 100 repeated cross-validation runs. In permutation analyses, the observed improvements in R$^2$, MAE, and RMSE all exceeded those obtained under 999 random reassignments of Canadian contextual information (all empirical permutation p = 0.001).

\subsection{Interpolating and Nowcasting Cardiovascular Disease Risk in US Counties}

\subsubsection{Data}

We obtained county-level cardiovascular disease (CVD) mortality counts for 2010--2023 from the National Vital Statistics System (NVSS) mortality files, aggregated by county of occurrence. CVD deaths were defined using ICD-10 codes I00--I99 as the underlying cause of death. Counts were harmonized to a consistent set of 3,091 county-equivalent units across the contiguous United States (48 states plus the District of Columbia), excluding Alaska, Hawaii, and Connecticut owing to PDFM embedding availability and county boundary changes during the study period. Counties with a total population below 500 in any year were also excluded. Two counties that underwent FIPS code changes (Shannon County, SD $\rightarrow$ Oglala Lakota County, FIPS 46113$\rightarrow$46102; Bedford City, VA consolidated into Bedford County, FIPS 51515$\rightarrow$51019) were remapped to their successor codes. County-level population denominators were drawn from the American Community Survey (ACS) 5-year estimates for the corresponding years.

County-level sociodemographic characteristics were obtained from the ACS 5-year estimates for 2009--2023. We extracted age distribution (percentage aged 25--44, 45--64, and $\ge$65 years), sex (percentage male), race and ethnicity (percentage non-Hispanic White, non-Hispanic Black, non-Hispanic American Indian/Alaska Native, and Hispanic), economic indicators (percentage of families in poverty, median household income in units of \$10,000, percentage unemployed), and population density (population per 1,000 per square mile, using Census Bureau land area files). All ACS covariates were standardized (zero mean, unit variance) across the analytic sample.

\subsubsection{Experimental Setup}

We evaluated model performance across two tasks:

\textbf{Nowcasting}. Models were trained on data from 2013--2022 (n = 3,091 counties $\times$ 10 years = 30,910 county-year observations) with 2023 outcome counts set to missing (Bayesian models) or held out as a test set (GBDT). Predicted counts for 2023 were compared against observed counts across all 3,091 counties. Because the training and test sets are fixed in this task, variability in GBDT performance was characterized by fitting 100 replicates that differed only in their random starting seed. Performance was averaged across replicates, with intervals reflecting the 2.5th and 97.5th percentiles.

\textbf{Spatial interpolation}. Using 2023 data only, counties were randomly partitioned into 5 folds (approximately 618 counties per fold). For each fold, models were trained on the 4 remaining folds and predicted counts for the held-out counties. Performance was averaged across folds, with intervals reflecting the lowest and highest fold-level estimates. For GBDT models, bootstrapping was used to generate 100 random hold-out samples, each fit with a distinct random starting seed. Replicates therefore varied in both the held-out counties and the stochastic components of model fitting. Performance was averaged across replicates, with intervals reflecting the 2.5th and 97.5th percentiles.

\subsubsection{Models}

\paragraph{Bayesian Spatiotemporal Models.}
We specified county-level CVD death counts using a negative binomial likelihood with an expected count offset, modeled via Integrated Nested Laplace Approximations (INLA)~\citep{bib31}. The expected count $E_{it}$ for each county\textsubscript{i}--year\textsubscript{t} observation was computed as $E_{it} = \frac{\sum y_{it}}{\sum N_{it}} \times N_{it}$ , where $y_{it}$ denotes observed counts and $N_{it}$ population sizes.

The general model form was:

\begin{equation}
\log(y_{it} / E_{it}) = \beta_0 + \mathbf{x}_{it}^{\top} \boldsymbol{\beta} + u_i + v_i + \gamma_t + \delta_t + \phi_{it}
\end{equation}

where:

\begin{itemize}
\item $E_{it}$ is the expected count for county $i$ at time $t$
\item $u_i + v_i$ is the BYM2 spatial random effect (a convolution of an intrinsic conditional autoregressive [ICAR] component $u_i$ and an unstructured component $v_i$, parameterized via the scaling of Riebler et al.)
\item $\gamma_t$ is a first-order random walk (RW1) for temporal trend
\item $\delta_t$ is an unstructured temporal random effect (IID)
\item $\phi_{it}$ is a space--time interaction term (IID)
\item $\mathbf{x}_{it}$ is the covariate vector, which varied across model specifications:
\begin{itemize}
\item No covariates: $\mathbf{x}_{it} = \emptyset$ (random effects only)
\item ACS, time-fixed: $\mathbf{x}_{it}$ = 10 standardized ACS variables fixed at 2023 values
\item PDFM: $\mathbf{x}_{it}$ = 330 standardized PDFM embedding dimensions from 2023
\item ACS + PDFM: both covariate sets
\end{itemize}
\end{itemize}

All random effects were assigned default INLA priors for the BYM2 specification~\citep{bib31,bib32} and negative binomial family. Models were fitted using R-INLA with R version 4.4.1. For each predicted observation we extracted the mean of the fitted Bayesian posterior distribution for each unit.

For the interpolation task, which used only 2023 cross-sectional data, the temporal random effects ($\gamma_t$, $\delta_t$) and space--time interaction ($\phi_{it}$) were omitted, and the model reduced to a spatial BYM2 specification.

For each task, the baseline Bayesian specification included no covariates. For nowcasting, the baseline comprised four components. The first was the expected-count offset, which reflects each county's population size and the overall background rate of CVD deaths. The second was the BYM2 county random effect, which combines an ICAR component for spatial dependence with an unstructured component for county heterogeneity that is not spatially dependent. The third was the temporal random effects (RW1 and IID), which capture non-linear temporal trends. The fourth was the county--year interaction random effect. For interpolation, the baseline comprised only the expected-count offset and the BYM2 county random effect. The covariate-augmented specifications added the ACS variables, the PDFM embeddings, or both to the corresponding baseline.

\paragraph{Gradient-boosted decision tree models.}
As an alternative modeling framework, we fit gradient-boosted decision trees (GBDT) using XGBoost with a squared-error objective. For the nowcasting task, the baseline feature set comprised total population and outcome counts lagged by one to five years. For the interpolation task, the baseline feature set comprised total population only. The covariate-augmented models added the ACS variables, the PDFM embeddings, or both to the corresponding baseline feature set. GBDT hyperparameters were chosen via validation RMSE and held constant across all specifications: maximum tree depth = 6, learning rate ($\eta$) = 0.05, subsample ratio = 0.8, column subsample per tree = 0.8, L2 regularization ($\lambda$) = 1, L1 regularization ($\alpha$) = 0, and 1,000 boosting rounds.

\subsubsection{Evaluation Metrics}

Performance was assessed using root mean squared error (RMSE) and mean absolute error (MAE).

Statistical significance was assessed by comparing each covariate-augmented model with its corresponding baseline model, as well as directly comparing ACS- and PDFM-augmented models, on paired county-level errors, using absolute errors for MAE and squared errors for RMSE. To account for spatial dependence, inference was clustered by state [n = 48 clusters]. Two criteria were applied. First, a clustered sign-flip permutation test randomly flipped the signs of state-level paired error differences to generate a null distribution of the mean difference. Second, a clustered bootstrap resampled states with replacement to obtain 95\% confidence intervals for the paired error differences. A difference was considered statistically significant only if the sign-flip test was significant and the bootstrap confidence interval excluded zero.

\subsection{Forecasting dengue in Mexico}

\subsubsection{Data}

Monthly dengue case counts for Mexico were compiled from national and subnational surveillance sources covering the period January 2015 through August 2025. Data from January 2020 onward were obtained from the Mexican Ministry of Health~\citep{bib33}, while earlier observations were assembled from historical state and local surveillance records. Records were harmonized and aggregated to GADM administrative level 2, yielding approximately 2,450 municipalities. For each municipality and month, we retained the observed number of dengue cases and the corresponding log-transformed case count.

\subsubsection{Experimental Setup}

We evaluated probabilistic forecasts of monthly dengue cases across approximately 2,450 Mexican municipalities (GADM Administrative Level 2 units) from January 2020 to August 2025. Forecasts were evaluated at 1-, 3-, and 6-month horizons.

A baseline TimesFM model generated probabilistic forecasts using historical dengue incidence together with the dynamic epidemiological and environmental covariates described in the Models section. PDFM embeddings were incorporated through a residual-learning framework, with separate residual models fitted for each forecast horizon and predictive quantile using a causal 24-month sliding training window. The predicted residuals were added to the corresponding TimesFM forecasts to produce the PDFM-adjusted predictive distributions.

To assess heterogeneity in forecast performance by local dengue case count, municipality-month forecasts were stratified retrospectively according to the observed number of cases in the forecast target month. Municipality-months with zero reported cases were treated as a separate stratum; within each target month, municipality-months with non-zero case counts were divided into three approximately equal-sized groups (terciles), representing low, medium and high observed case counts.

Forecast performance was evaluated using the Weighted Interval Score (WIS), with lower values indicating better probabilistic forecast performance. We report differences as $\Delta$WIS = WIS\textsubscript{PDFM-adjusted} $-$ WIS\textsubscript{TimesFM}, such that negative values indicate improved performance after incorporating PDFM embeddings. As a sensitivity analysis, we repeated the analysis using PDFM embeddings from October 2025, holding all other modeling and evaluation procedures fixed.

\subsubsection{Models}

We evaluated probabilistic dengue forecasts using TimesFM 2.0 (\nolinkurl{google/timesfm-2.0-500m-pytorch})~\citep{bib17} configured in ``xreg + timesfm'' mode with monthly frequency (freq = 1). Forecasts were generated at 1-, 3-, and 6-month horizons.

The baseline forecasting model incorporated a set of dynamic epidemiologically relevant covariates alongside historical dengue incidence. Covariates included lagged dengue incidence (log cases at lag 1), minimum temperature, maximum temperature, and precipitation (lags 0, 1, and 2), the six-month Standardized Precipitation Index (SPI6; lag 0), the Oceanic Ni\~no Index (ONI; lag 0 together with 6- and 12-month averages), and population size derived from WorldPop~\citep{bib34}. Climate variables were obtained from WorldClim~\citep{bib35}. Dynamic numerical covariates were supplied directly to TimesFM through the forecast\_with\_covariates framework.

To evaluate the contribution of geospatial foundation model covariates, PDFM embeddings were incorporated through a residual-learning framework. For each forecast horizon and predictive quantile, residuals between the observed dengue incidence and the corresponding TimesFM forecast were modeled using a histogram-based gradient boosting quantile regression model (HistGradientBoostingRegressor, scikit-learn), with PDFM embedding features as predictors. Models were refitted causally at each forecast target date using a 24-month sliding training window and quantile (pinball) loss. Each model used 200 boosting iterations, a learning rate of 0.1, and a maximum of 31 leaf nodes. Predicted residuals were added to the corresponding TimesFM quantile forecasts to produce the final PDFM-adjusted predictive distributions.

\subsubsection{Evaluation Metrics}

Forecast performance was evaluated using the Weighted Interval Score (WIS), a proper scoring rule for probabilistic forecasts that jointly assesses both calibration and sharpness of the predictive distribution~\citep{bib36}. WIS was calculated from the set of forecast quantiles and averaged across municipalities and forecast dates. Lower WIS values indicate improved probabilistic forecast performance. Results are reported as $\Delta$WIS = WIS\textsubscript{PDFM-adjusted} $-$ WIS\textsubscript{TimesFM}, such that negative values indicate improved probabilistic forecast performance following PDFM adjustment.

Statistical differences in predictive accuracy were evaluated using two-sided Diebold--Mariano tests~\citep{bib37} on chronological target-month WIS sequences. To account for temporal dependence, 95\% confidence intervals for $\Delta$WIS were estimated using a temporal moving-block bootstrap~\citep{bib38} with three-month blocks; sensitivity analyses used six- and twelve-month blocks.

\subsection{Cholera Hotspot Forecasting}

\subsubsection{Data}

For this particular use case, low internet penetration and sparse digital coverage in the DRC yield too few observations per spatial unit to train the self-supervised encoder described above (see PDFM Embeddings). We therefore used a custom, low-resource version of the embeddings. It draws on the same three input modalities as the standard model: aggregated search trends, the built environment (POI density and aggregated place-visit counts), and environmental determinants (meteorology and air quality). It differs in two respects. First, to offset sparse search activity, aggregated search counts were pooled over the six months ending in October 2023, rather than the single month used for standard PDFM embeddings. Second, the small number of spatial units and the sparsity of the inputs ruled out the self-supervised reconstruction framework. The input features were instead compressed to 200 dimensions using principal component analysis (PCA). As in the other case studies, they were used as a single static snapshot for every health-zone-week. Because all their inputs predate the 1 November 2023 evaluation boundary, no evaluation-period data informed their construction.

Weekly aggregated cholera surveillance records for the DRC were obtained from the national integrated disease surveillance system, covering epidemiological week 1 of 2010 through week 29 of 2025. Records were processed into a complete, single-valued health-zone $\times$ week panel as follows.

Each record's (year, epidemiological week) pair was converted to the Monday of the corresponding ISO week. For records assigned to week 53 of a year that has no ISO week 53, the date was set to 28 December of that year, which always falls in the final ISO week. Records whose year--week pair could not be resolved to a valid date were excluded.

Health zone names in the surveillance system were entered as free text and varied in case, spacing, and punctuation. Names were therefore standardized by converting to upper case, trimming surrounding whitespace, replacing hyphens, apostrophes and backticks with spaces, and collapsing repeated whitespace. A curated synonym dictionary was then applied to reconcile remaining variants against the canonical health zone gazetteer --- for example mapping ``BAGIRA KASHA'' to ``BAGIRA'', ``LULENGE KIMBI'' to ``KIMBI LULENGE'', and ``SHABUNDA CENTRE'' to ``SHABUNDA''. This consolidation yielded 403 canonical health zones.

The surveillance system contains multiple records for the same health zone and week, arising from resubmission and retrospective correction. Where duplicates disagreed on the case count, the most frequently reported value was retained, with ties broken toward the larger count; a single record per health-zone-week was kept.

Health zones did not report every week, and a non-reporting week was not distinguishable in the source data from a week with no cases. We constructed the complete cross product of all 403 health zones with all 811 observed weeks and left-joined the deduplicated records onto it, yielding 326,833 health-zone-week observations. Case and death counts for health-zone-weeks with no corresponding record were set to zero, as were missing counts within existing records.

Reported populations below 10,000 were treated as data-entry errors, set to missing, and imputed within each health zone by forward-fill followed by backward-fill, so that every health-zone-week carries a population value.

Health zones were assigned to endemicity regimes using only surveillance data preceding the evaluation boundary. Using definitions discussed in prior work, a zone was classified as endemic if at least 156 weeks of surveillance were available before the boundary and cholera cases were reported in at least 50\% of those weeks; dormant if no cases were ever reported before the boundary; and sporadic otherwise~\citep{bib24}. This yielded 15 endemic, 340 sporadic and 48 dormant zones. Restricting regime assignment to pre-boundary data ensures that stratum membership cannot be informed by the outcomes being predicted. The dormant stratum recorded no emergence events during the evaluation period at any lead time, leaving PR-AUC undefined and every weekly Precision@5 comparison an exact tie; it was therefore excluded from the stratified analysis and from the associated multiple-comparison family.

Two feature sets were compared.

The History-only set comprised twelve covariates derived exclusively from information available at or before week t: suspected case counts at lags of 1, 2 and 4 weeks; the cumulative historical maximum and cumulative historical total of weekly cases, each lagged by one week; rolling 12-, 26- and 52-week case sums, each lagged by one week; the number of weeks since the zone last reported a non-zero case count; log-transformed population; and sine and cosine transforms of the epidemiological week number to represent annual seasonality. All rolling and cumulative features were shifted by one week before use to preclude same-week leakage.

The History+PDFM set comprised these twelve covariates augmented with the PDFM embeddings.

\subsubsection{Models}

For each lead time we fitted a histogram-based gradient-boosted decision tree classifier (scikit-learn HistGradientBoostingClassifier; 100 boosting iterations, learning rate 0.05, maximum 20 leaf nodes, minimum 30 samples per leaf, L2 regularization 2.0). Class imbalance was addressed by weighting positive training examples by the ratio of negative to positive training examples.

Evaluation was strictly prospective with a zero-lookahead boundary at 1 November 2023. To prevent label leakage across the boundary, the training set for lead time h was truncated at h weeks before the boundary, so that no training label could depend on surveillance observed during the evaluation period. The evaluation set comprised all eligible zone-weeks from 1 November 2023 to 14 July 2025. Models were never refitted during the evaluation period, and no evaluation-period data entered feature construction, dimensionality selection, or regime assignment.

To reduce sensitivity to the stochastic components of tree fitting, each configuration was fitted with five random seeds and the predicted probabilities were averaged across seeds before evaluation. All reported statistics are computed on these seed-averaged predictions.

\subsubsection{Evaluation Metrics}

We forecast emergence: the transition of a health zone from a non-outbreak state into an outbreak state within a specified lead time. An outbreak onset was defined as a three-week window in which the cumulative suspected case count reached 100 or more (T2\_Sum100)~\citep{bib24}.

For a lead time of h weeks, a health-zone-week t was eligible for evaluation if the zone was in a non-outbreak state at t (fewer than 50 cases that week), at least two weeks of prior surveillance were available, and at least h weeks of subsequent surveillance were available. An eligible zone-week was labeled positive if any outbreak onset window began in weeks t+1 through t+h inclusive, and negative otherwise. Ineligible zone-weeks were excluded from both training and evaluation, so the models were never asked to predict emergence for a zone already in an outbreak state. Labels were generated independently for each lead time.

PR-AUC was computed as the average precision over all eligible zone-weeks at a given lead time, pooling across the evaluation period. Average precision was chosen over the area under the receiver operating characteristic curve because emergence events are rare (0.69--23.4\% of eligible zone-weeks depending on stratum and lead time) and precision-recall summaries are more informative than ROC-based summaries under severe class imbalance.

Precision@5 was computed weekly to reflect operational shortlisting: within each evaluation week, health zones were ranked by predicted probability, ties broken against the positive class, and the proportion of true emergence events among the top five zones was recorded. The reported value is the mean across evaluation weeks. Weeks with fewer than five eligible zones were excluded. Precision@5 was reported only for strata containing at least ten health zones, since a top-five list drawn from fewer than twice that many candidates degenerates toward the stratum base rate.

All inference compares History+PDFM against History-only on identical evaluation rows, so all tests are paired. For PR-AUC, we used a paired block bootstrap over evaluation weeks with 2,000 resamples. Whole weeks were resampled with replacement to preserve within-week dependence between health zones; both models were re-scored on each resampled set and the difference in average precision retained. Two-sided P values were obtained from the proportion of resamples on either side of zero, and 95\% confidence intervals from the 2.5th and 97.5th percentiles of the resampled difference distribution. For Precision@5, the unit of analysis is the evaluation week, and the paired weekly differences are heavily tied, since the two models frequently produce shortlists with identical event counts. We therefore used a paired sign-flip permutation test with 50,000 permutations: the sign of each weekly difference was randomly flipped to generate the null distribution of the mean difference, and the two-sided P value computed with the Phipson--Smyth correction, which is exact under the null and requires no convention for handling ties. Confidence intervals for Precision@5 differences were obtained by percentile bootstrap over evaluation weeks with 2,000 resamples. We additionally report the number of weeks in which each model produced the better shortlist.

Multiple comparisons were controlled using the Benjamini--Hochberg procedure at a false discovery rate of 5\%. Correction families were defined per table: the unstratified analysis was corrected over its eight tests (two metrics $\times$ four lead times), and the stratified analysis over its sixteen tests (two regimes $\times$ two metrics $\times$ four lead times). Adjusted values are reported as q. Because the endemicity stratification was specified after the pooled analysis, the stratified family is corrected in full --- including comparisons that were not of primary interest --- rather than only over the comparisons reported as significant.

All random seeds were fixed and all analyses are deterministic given the seed-averaged predictions.

\subsection{Postpartum Depression Screening (PRAMS)}

\subsubsection{Data}

Cross-sectional data from the CDC Pregnancy Risk Assessment Monitoring System (PRAMS), Phases 7--8 (births 2012--2021), obtained under a CDC data-use agreement. Of 390,320 respondents, 332,970 were included (Extended Data Table~\ref{exttab6}):

\begin{itemize}
\item both depressive-symptom items answered ($-$9,631);
\item core predictors complete ($-$24,063);
\item NCHS urban/rural classification known ($-$5,903);
\item jurisdiction covered by the PDFM release used ($-$17,753; Alaska and Hawaii absent).
\end{itemize}

The primary outcome was the CDC postpartum depressive-symptoms indicator: positive when either MH\_PPDPR (``Since your new baby was born, how often have you felt down, depressed, or hopeless?'') or MH\_PPINT (little interest or pleasure in doing things) was answered ``always'' or ``often''. The highest-severity response (MH\_PPDPR ``always'') was a pre-registered secondary outcome.

Predictors were state of residence, maternal education, smoking during pregnancy, pre-pregnancy BMI and marital status. Socioeconomic comparisons added household income as a percentage of the federal poverty level and Medicaid-paid delivery. Race and ethnicity were excluded from the feature set by design.

PDFM provides 330-dimensional embeddings for US ZIP Code Tabulation Areas (ZCTAs); this analysis used the first public release (v0). The PRAMS research file locates respondents only by state and by county-based urban/rural status (NCHS classification). ZCTA embeddings were therefore mapped to counties (largest land-area overlap, Census 2020) and to NCHS 2013 urban/rural codes~\citep{bib39}, then averaged with population weights to state $\times$ urban/rural locations (87 locations).

Each respondent received exactly one embedding (a one-to-one join), which avoids the duplication of respondents that a many-to-one linkage causes. A state-level embedding, random embeddings and shuffled embeddings served as comparators.

\subsubsection{Models}

We trained gradient-boosted decision trees (LightGBM) and elastic-net logistic regression to predict the binary CDC postpartum depressive-symptoms indicator: MH\_PPDPR or MH\_PPINT answered ``always'' or ``often'' (prevalence 13.95\%; 12.5\% survey-weighted). The baseline model used the individual-level predictors listed under Data. Race and ethnicity were excluded by design. The PDFM model replaced state with the population-weighted PDFM embedding of each respondent's state $\times$ urban/rural location (87 locations).

Comparator models added:

\begin{itemize}
\item an urban/rural flag;
\item location indicators without PDFM;
\item individual income (as \% of the federal poverty level) and Medicaid status;
\item PDFM on top of income and Medicaid;
\item an area poverty share computed within the training fold.
\end{itemize}

Random- and shuffled-embedding controls tested whether gains came from PDFM's content.

Hyperparameters were tuned by nested random search with the same budget for every model (40 trials per learner; objective log-loss, a proper scoring rule). The learner was chosen by inner cross-validation on training data only. No class weights were used, so predicted probabilities remain calibrated. All preprocessing was fitted inside training folds.

Test sets were drawn once and sealed before any model was fitted:

\begin{itemize}
\item A 20\% respondent holdout (S1; n = 66,594; primary);
\item Ten folds that each hold out whole jurisdictions (S2);
\item Births in 2020--2021, with models trained on 2012--2019 (S3).
\end{itemize}

The analysis plan was pre-registered before any outcome was modeled.

\subsubsection{Evaluation Metrics}

Primary metrics were threshold-free: AUC, Brier score and calibration (observed/expected ratio and calibration slope), reported unweighted and weighted by the PRAMS survey weights (Extended Data Table~\ref{exttab3}). To compare screening performance without tuning thresholds on test data, screening was compared at a fixed screening capacity: the 20\% of mothers with the highest predicted risk were flagged, a share fixed in advance, with 10\% and 30\% as sensitivity analyses. We report sensitivity, positive predictive value and decision-curve net benefit over risk thresholds of 5--30\%.

Three comparisons were pre-specified, with Holm correction (Extended Data Table~\ref{exttab4}):

\begin{itemize}
\item PDFM vs baseline;
\item Whether PDFM can replace individual income and Medicaid (non-inferiority margin $-$0.005 AUC, plus the share of their gain recovered);
\item Whether PDFM adds to them.
\end{itemize}

Holm's step-down procedure~\citep{bib40} controlled the family-wise error rate across these three comparisons in S1, the primary setting. A comparison was claimed only if its Holm-adjusted p-value was below 0.05 and its confidence interval supported the pre-defined reading. P-values came from the paired bootstrap, as (count + 1)/(B + 1) with B = 1,000: two-sided for the first and third comparisons, and one-sided against the $-$0.005 margin for non-inferiority. Results in other settings and all secondary analyses are reported without adjustment.

95\% confidence intervals came from 1,000 paired bootstrap resamples, resampling whole jurisdictions in S2.

Fairness was assessed with a two-stage audit of race $\times$ ethnicity $\times$ urban/rural groups. Performance gaps were identified on development data and then tested once on the sealed test sets. A gap counted as closed only if it narrowed with a confidence interval excluding zero and no other group lost more than 2 percentage points of sensitivity. Race and ethnicity defined evaluation groups only and were never model inputs.

Controls: label permutation; random and shuffled embeddings; and a positive control that predicts area poverty from the embeddings.

Exploratory analyses (labeled; Benjamini--Hochberg-adjusted where multiple groups were compared):

\begin{itemize}
\item An 80\%-sensitivity threshold derived on training folds only
\item Youden's J
\item Intersectional groups at three levels
\item \textit{Projections to US and global births}, including a screening simulator: test-set predictions were reweighted to each target setting's rural share and PPD prevalence (keeping the observed score distributions within urban/rural $\times$ PPD strata and the US rural:urban prevalence ratio), and the screening rule was re-applied within the reweighted population. The simulator was verified by reproducing the US estimates exactly at US composition.
\end{itemize}

\textit{Screening simulator}. To ask what the S2 effects would mean in settings with a different rural share and PPD prevalence, we wrote the screening process as executable code and verified it against the observed data, following the idea of verified code world models~\citep{bib25}. The simulator takes the pooled S2 test predictions (jurisdictions unseen in training) and rescales the survey weights so that the target population has the target rural share and overall prevalence. Within each urban/rural $\times$ PPD stratum, the distribution of model scores is kept as observed, and the US rural:urban prevalence ratio is preserved. The screening rule is then re-run inside the target population: the fixed capacity (10\%, 20\% or 30\% of mothers flagged) is re-selected under the target weights, while the 80\%-sensitivity thresholds, set on training folds, stay fixed. Outputs are changes in cases found and in false alarms per 100,000 births, with 95\% CIs from 1,000 bootstrap resamples of jurisdictions. Verification is a unit test: run at US composition, the simulator must reproduce the US estimates exactly. Each scenario paired a region's rural share~\citep{bib41,bib42} with its pooled prevalence~\citep{bib21}; two global prevalence estimates~\citep{bib21,bib22} and low-income countries were added, and all scenarios are reported (Extended Data Table~\ref{exttab5}). The scenarios were defined after the primary results were known and are exploratory. They assume that US score distributions carry over to other settings; because PDFM v0 and PRAMS are US-only, they cannot show that PDFM works elsewhere.

Ethics statement: PRAMS data were obtained from CDC under a data-use agreement, which CDC granted after reviewing the proposed analysis. Only de-identified data were analyzed; no institutional review board review was sought.

Reporting follows TRIPOD+AI~\citep{bib43}.

\subsection{Use of AI in Manuscript Preparation}

Portions of the manuscript were drafted manually and then further refined using LLMs. Final manual checks were performed to ensure content accuracy. The authors take full responsibility for the content.

\section*{Data Availability}
Google offers a research access program for academic and non-commercial investigators to access Population Dynamics Foundation Model (PDFM) embeddings. Qualified researchers can request access by contacting pdfm-embeddings@google.com. Public code repositories, tutorials, and benchmark replication examples are available at \url{https://github.com/google-research/population-dynamics}.

All non-Google empirical health datasets with the exception of the cholera dataset were obtained, curated, and analyzed independently by the respective academic and institutional collaborators, with Google not accessing protected underlying individual or patient-level records:

\textbf{Measles--Mumps--Rubella (MMR) Vaccination Coverage:} County-level estimates of pediatric MMR vaccination uptake were generated by researchers at Boston Children's Hospital, Harvard Medical School, and the Icahn School of Medicine at Mount Sinai using digital participatory surveillance data from the OutbreaksNearMe (ONM) platform. Canadian spatial context was derived using Statistics Canada Forward Sortation Area (FSA) boundary files, publicly available via Statistics Canada Catalogue no. 92-179-X (\url{https://www150.statcan.gc.ca/n1/en/catalogue/92-179-X}).

\textbf{Cardiovascular Disease Mortality:} County-level cardiovascular disease mortality counts (ICD-10 codes I00--I99) across 2010--2023 were derived from the National Vital Statistics System (NVSS) microdata files curated by the National Center for Health Statistics (NCHS) of the Centers for Disease Control and Prevention (CDC). Aggregate tabulations are accessible through CDC WONDER (\url{https://wonder.cdc.gov/mcd.html}) queries, with suppression of counts < 10. Researchers seeking unsuppressed subnational tabulations must apply for a data use agreement pursuant to the NCHS Data Release Policy (\url{https://www.cdc.gov/nchs/nvss/dvs_data_release.htm}). County socioeconomic, demographic, and population covariates were retrieved from the US Census Bureau American Community Survey (ACS) 5-year estimates (2009--2023) via data.census.gov or the Census API (\url{https://www.census.gov/programs-surveys/acs.html}).

\textbf{Municipal Dengue Surveillance:} Monthly municipal dengue case counts from January 2020 onward were obtained directly from the Mexican Ministry of Health (Secretar\'{\i}a de Salud) open-data portal (gob.mx/salud). Probable and confirmed dengue cases were retained and assigned temporally using the reported date of symptom onset; discarded cases and records without valid Mexican municipality identifiers were excluded. Data from 2015--2019 were provided by the Directorate of the National Institute of Epidemiological Reference of the Mexican Ministry of Health. Records from both sources were harmonized to common municipality-level units for analysis.

\textbf{Cholera Emergence Forecasting:} Epidemiological surveillance records of suspected cholera cases and deaths across 403 health zones (\textit{zones de sant\'e}) in the DRC (2010--2025) were compiled through the National Integrated Disease Surveillance and Response System (IDSRS) by the DRC Ministry of Public Health, Hygiene and Prevention (Direction de Surveillance \'Epid\'emiologique), in collaboration with the World Health Organization Regional Office for Africa (WHO AFRO). Access to disaggregated subnational weekly surveillance data is governed by national data sharing frameworks. Regional outbreak summaries are published in the WHO AFRO Outbreaks and Other Emergencies bulletins (\url{https://www.afro.who.int/health-topics/cholera}).

\textbf{Postpartum Depression Screening:} Individual-level responses from the Pregnancy Risk Assessment Monitoring System (PRAMS), Phases 7 and 8 (births 2012--2021), were obtained from the Centers for Disease Control and Prevention (CDC) under a data-use agreement that prohibits redistribution by the authors. Researchers may request the same de-identified multi-jurisdiction data by submitting a proposal through the CDC PRAMS Automated Research File (ARF) system or to PRAMSProposals@cdc.gov (\url{https://www.cdc.gov/prams/php/data-research/index.html}). PDFM ZIP Code Tabulation Area embeddings (v0 release) are publicly available at \url{https://github.com/google-research/population-dynamics}. The US Census Bureau 2020 ZCTA-to-county relationship file and the NCHS 2013 Urban--Rural Classification Scheme for Counties are publicly available.

\section*{Code Availability}
The pre-training code for the Population Dynamics Foundation Model (PDFM) is proprietary and not publicly available; public tutorials, starter code, and benchmark replication examples for using PDFM embeddings are available at \url{https://github.com/google-research/population-dynamics}. Downstream analyses for each case study were performed independently at the respective collaborating institutions and relied exclusively on standard, off-the-shelf open-source software and libraries---including Python (scikit-learn, XGBoost, LightGBM, PyTorch, and TimesFM 2.0) and R (v.4.4.1, R-INLA)---rather than custom algorithms. Because these downstream scripts are tied to institution-specific data pipelines and restricted public health datasets, they are not publicly released but are available from the corresponding authors, and the respective case-study leads, upon reasonable request, including for peer review; full mathematical specifications, feature definitions, and hyperparameter settings are provided in the Methods to enable independent replication.

\section*{Author Contributions}
A.M., J. Bien, H. Sadeghi, M.S., J.P., Y.L., T.S., C.E., S.S. and G.P. developed the foundation model and core methodology. Domain case studies, including data acquisition, analysis and supervision, were led as follows: MMR vaccination coverage (E.Z., J. Brownstein and B.R.); cardiovascular disease mortality (K.W.-M., S.C., B.S., D.N., L.T. and M.G.); municipal dengue forecasting (J. Brittain, J. Bien, E.L.O., J.A.D.-Q. and M.U.G.K.); cholera hotspot forecasting (A.M., H. Sadeghi, S.J., C.E., I.F.C. and E.L.K.); and postpartum depression risk stratification (M.F.). A.C.-Y.C., A.C. and K.W. contributed to paper organization and team logistics. A.B., R.S., A.H., N.E., Y.M., K.W., M.B., S.S. and G.P. provided leadership and strategic guidance. A.M., A.S., K.W.-M., E.Z., J. Brittain, M.F., H. Subramaniam, J. Bien, S.C., Y.L., A.C.-Y.C., M.U.G.K., I.F.C., E.L.K., M.B., B.R., L.T., M.G., S.S. and G.P. contributed to drafting and revising the manuscript.

\section*{Competing Interests}
A.M., A.S., H. Subramaniam, J. Bien, H. Sadeghi, S.J., M.S., J.P., Y.L., T.S., A.C.-Y.C., A.C., C.E., A.B., R.S., A.H., N.E., Y.M., K.W., M.B., S.S. and G.P. are employees of Google LLC and own Alphabet stock as part of the standard compensation package. Google develops the Population Dynamics Foundation Model evaluated in this study. M.F. is supported by a Google Health grant for wearable monitoring of mental health in pregnancy; that work did not fund the analyses reported here. The other authors declare no competing interests.

\bibliography{main}

\clearpage
\appendix
\renewcommand{\tablename}{Extended Data Table}
\setcounter{table}{0}

\section*{Extended Data}

\begin{table}[htbp]
\caption{Cumulative population in U.S. border counties affected by incorporating Canadian contextual information, by threshold of change in predicted MMR vaccination coverage.}\label{exttab1}%
\footnotesize
\setlength{\tabcolsep}{0pt}
\begin{tabular*}{\textwidth}{@{\extracolsep\fill}>{ \centering\arraybackslash }p{0.14\textwidth}>{ \centering\arraybackslash }p{0.13\textwidth}>{ \centering\arraybackslash }p{0.15\textwidth}>{ \centering\arraybackslash }p{0.16\textwidth}>{ \centering\arraybackslash }p{0.14\textwidth}>{ \centering\arraybackslash }p{0.13\textwidth}@{}}
\toprule
\textbf{Min. change (pp)} & \textbf{Border counties} & \textbf{\% Border counties} & \textbf{Affected population} & \textbf{\% Border pop.} & \textbf{\% Total pop.} \\
\midrule
10.0 & 3 & 2.1 & 241,045 & 1.3 & 0.1 \\
7.5 & 9 & 6.2 & 1,672,517 & 9.0 & 0.5 \\
5.0 & 24 & 16.4 & 2,502,339 & 13.4 & 0.8 \\
4.0 & 35 & 24.0 & 3,529,624 & 19.0 & 1.1 \\
3.0 & 55 & 37.7 & 4,706,150 & 25.3 & 1.5 \\
2.0 & 74 & 50.7 & 8,540,213 & 45.9 & 2.6 \\
1.0 & 108 & 74.0 & 13,120,238 & 70.5 & 4.0 \\
\botrule
\end{tabular*}
\end{table}

\begin{table}[htbp]
\caption{Performance stratified by pre-evaluation endemicity regime}\label{exttab2}%
\footnotesize
\setlength{\tabcolsep}{0pt}
\begin{tabular*}{\textwidth}{@{\extracolsep\fill}lllccccc@{}}
\toprule
\textbf{Regime} & \textbf{Lead Time} & \textbf{Metric} & \textbf{History-only} & \textbf{History+PDFM} & \textbf{$\Delta$ (\%)} & \textbf{95\% CI} & \textbf{\textit{p}} \\
\midrule
Endemic & 1 wk & PR-AUC & 0.4679 & 0.4748 & $+$0.0069 ($+$1.5) & $-$0.0388 to $+$0.0494 & 0.99 \\
Endemic & 1 wk & Precision@5 & 0.2318 & 0.2250 & $-$0.0068 ($-$2.9) & $-$0.0159 to $+$0.0023 & 0.74 \\
Endemic & 2 wk & PR-AUC & 0.4637 & 0.4444 & $-$0.0193 ($-$4.2) & $-$0.0611 to $+$0.0251 & 0.75 \\
Endemic & 2 wk & Precision@5 & 0.2621 & 0.2644 & $+$0.0023 ($+$0.9) & $-$0.0046 to $+$0.0115 & 1.00 \\
Endemic & 4 wk & PR-AUC & 0.4209 & 0.4729 & $+$0.0521 ($+$12.4) & $+$0.0172 to $+$0.0887 & 0.024 \\
Endemic & 4 wk & Precision@5 & 0.3082 & 0.3200 & $+$0.0118 ($+$3.8) & $-$0.0024 to $+$0.0259 & 0.47 \\
Endemic & 8 wk & PR-AUC & 0.4143 & 0.4449 & $+$0.0306 ($+$7.4) & $-$0.0097 to $+$0.0723 & 0.40 \\
Endemic & 8 wk & Precision@5 & 0.3333 & 0.3975 & $+$0.0642 ($+$19.3) & $+$0.0370 to $+$0.0938 & 0.0010 \\
\midrule
Sporadic & 1 wk & PR-AUC & 0.1806 & 0.1817 & $+$0.0011 ($+$0.6) & $-$0.0161 to $+$0.0185 & 1.00 \\
Sporadic & 1 wk & Precision@5 & 0.1705 & 0.1636 & $-$0.0068 ($-$4.0) & $-$0.0250 to $+$0.0114 & 0.99 \\
Sporadic & 2 wk & PR-AUC & 0.2046 & 0.1944 & $-$0.0102 ($-$5.0) & $-$0.0286 to $+$0.0074 & 0.55 \\
Sporadic & 2 wk & Precision@5 & 0.1977 & 0.1954 & $-$0.0023 ($-$1.2) & $-$0.0207 to $+$0.0161 & 1.00 \\
Sporadic & 4 wk & PR-AUC & 0.2127 & 0.2397 & $+$0.0270 ($+$12.7) & $+$0.0067 to $+$0.0470 & 0.059 \\
Sporadic & 4 wk & Precision@5 & 0.2282 & 0.2541 & $+$0.0259 ($+$11.3) & $-$0.0024 to $+$0.0541 & 0.40 \\
Sporadic & 8 wk & PR-AUC & 0.2085 & 0.2059 & $-$0.0026 ($-$1.3) & $-$0.0232 to $+$0.0165 & 0.99 \\
Sporadic & 8 wk & Precision@5 & 0.2444 & 0.2543 & $+$0.0099 ($+$4.0) & $-$0.0247 to $+$0.0444 & 0.99 \\
\botrule
\end{tabular*}
\vspace{4pt}\par\noindent{\footnotesize \textit{Regimes assigned from surveillance data preceding 1 November 2023 only: Endemic ($\ge$156 weeks surveillance and $\ge$50\% of weeks having reported cases; n = 15 zones) and Sporadic (n = 340 zones). Dormant zones (n = 48) recorded zero events during evaluation and are omitted. p values are Benjamini--Hochberg adjusted across the table.}}
\end{table}

\begin{table}[htbp]
\caption{Test-set performance (unweighted)}\label{exttab3}%
\footnotesize
\setlength{\tabcolsep}{0pt}
\begin{tabular*}{\textwidth}{@{\extracolsep\fill}>{ \raggedright\arraybackslash }p{0.18\textwidth}>{ \raggedright\arraybackslash }p{0.24\textwidth}cc>{ \centering\arraybackslash }p{0.14\textwidth}>{ \centering\arraybackslash }p{0.13\textwidth}@{}}
\toprule
\textbf{Setting} & \textbf{Model} & \textbf{AUC (95\% CI)} & \textbf{Brier score} & \textbf{Observed / expected} & \textbf{Calibration slope} \\
\midrule
S1: seen states (primary) & Baseline & 0.622 (0.616--0.629) & 0.1174 & 1.00 & 1.05 \\
 & PDFM (replaces state) & 0.624 (0.618--0.630) & 0.1173 & 1.00 & 1.02 \\
 & Baseline + income, Medicaid & 0.636 (0.630--0.642) & 0.1168 & 1.00 & 1.06 \\
 & Income, Medicaid + PDFM & 0.637 (0.631--0.643) & 0.1167 & 1.00 & 1.10 \\
\midrule
S2: unseen states & Baseline & 0.615 (0.604--0.625) & 0.1177 & 1.00 & 1.02 \\
 & PDFM (replaces state) & 0.618 (0.607--0.630) & 0.1176 & 1.01 & 1.06 \\
 & Baseline + income, Medicaid & 0.629 (0.619--0.638) & 0.1171 & 1.00 & 1.03 \\
 & Income, Medicaid + PDFM & 0.631 (0.621--0.641) & 0.1170 & 1.00 & 1.06 \\
\midrule
S3: 2020--21 births & Baseline & 0.619 (0.613--0.625) & 0.1195 & 1.04 & 0.94 \\
 & PDFM (replaces state) & 0.620 (0.613--0.625) & 0.1194 & 1.04 & 0.95 \\
 & Baseline + income, Medicaid & 0.629 (0.623--0.635) & 0.1190 & 1.06 & 0.97 \\
 & Income, Medicaid + PDFM & 0.630 (0.624--0.636) & 0.1190 & 1.06 & 0.95 \\
\botrule
\end{tabular*}
\vspace{4pt}\par\noindent{\footnotesize \textit{Weighted observed/expected ratios in S1 are 0.92--0.93. PDFM = population-weighted PDFM v0 embedding of the respondent's state $\times$ urban/rural location.}}
\end{table}

\begin{table}[htbp]
\caption{Pre-registered comparisons ($\Delta$AUC unless stated)}\label{exttab4}%
\footnotesize
\setlength{\tabcolsep}{0pt}
\begin{tabular*}{\textwidth}{@{\extracolsep\fill}>{ \raggedright\arraybackslash }p{0.18\textwidth}>{ \raggedright\arraybackslash }p{0.32\textwidth}>{ \raggedright\arraybackslash }p{0.23\textwidth}>{ \raggedright\arraybackslash }p{0.13\textwidth}c@{}}
\toprule
\textbf{Setting} & \textbf{Comparison} & \textbf{Estimate (95\% CI)} & \textbf{Pre-defined reading} & \textbf{Holm \textit{p} (S1)} \\
\midrule
S1: seen states (primary) & PDFM vs baseline & $+$0.0020 ($+$0.0008 to $+$0.0031) & improvement & 0.012 \\
 & PDFM instead of income and Medicaid (non-inferiority, margin $-$0.005) & $-$0.0116 ($-$0.0144 to $-$0.0089) & harm & 1.000 \\
 & Share of the income/Medicaid gain recovered by PDFM ($R$) & 0.15 (0.06 to 0.24) & $R$ not above 0.5 & --- \\
 & PDFM vs area poverty share & $+$0.0022 ($+$0.0011 to $+$0.0033) & improvement & --- \\
 & PDFM added to income and Medicaid & $+$0.0011 ($+$0.0000 to $+$0.0022) & improvement & 0.104 \\
\midrule
S2: unseen states & PDFM vs baseline & $+$0.0038 ($+$0.0004 to $+$0.0072) & improvement & --- \\
 & PDFM instead of income and Medicaid (non-inferiority, margin $-$0.005) & $-$0.0102 ($-$0.0136 to $-$0.0064) & harm & --- \\
 & Share of the income/Medicaid gain recovered by PDFM ($R$) & 0.27 (0.03 to 0.52) & $R$ not above 0.5 & --- \\
 & PDFM added to income and Medicaid & $+$0.0020 ($-$0.0008 to $+$0.0049) & no effect & --- \\
\midrule
S3: 2020--21 births & PDFM vs baseline & $+$0.0006 ($-$0.0005 to $+$0.0015) & no effect & --- \\
 & PDFM instead of income and Medicaid (non-inferiority, margin $-$0.005) & $-$0.0098 ($-$0.0126 to $-$0.0068) & harm & --- \\
 & Share of the income/Medicaid gain recovered by PDFM ($R$) & 0.05 ($-$0.05 to 0.15) & $R$ not above 0.5 & --- \\
 & PDFM vs area poverty share & $+$0.0014 ($+$0.0005 to $+$0.0023) & improvement & --- \\
 & PDFM added to income and Medicaid & $+$0.0010 ($+$0.0000 to $+$0.0021) & improvement & --- \\
\botrule
\end{tabular*}
\vspace{4pt}\par\noindent{\footnotesize \textit{Under the three pre-registered S1 comparisons, ``PDFM added to income and Medicaid'' (lower CI bound 0.0000; Holm-adjusted p = 0.104) is not claimed. Readings follow the pre-registered rules (improvement = CI above 0; no effect = CI includes 0 and $|\Delta| < 0.01$). None of 21 pre-registered intersectional gaps closed in S1 or S2. The pre-registered label-permutation control gave AUC 0.475 (0.468--0.481); a post-hoc null of 20 permutations centered on 0.499 (0.479--0.522). Holm's step-down procedure}~\citep{bib40} \textit{controlled the family-wise error rate across these three comparisons in S1, the primary setting. A comparison was claimed only if its Holm-adjusted p-value was below 0.05 and its confidence interval supported the pre-defined reading. P-values came from the paired bootstrap, as (count + 1)/(B + 1) with B = 1,000: two-sided for the first and third comparisons, and one-sided against the $-$0.005 margin for non-inferiority. Results in other settings and all secondary analyses are reported without adjustment.}}
\end{table}

\begin{table}[htbp]
\caption{Screening simulator: PDFM vs baseline transported to other settings (exploratory)}\label{exttab5}%
\fontsize{7.5}{9}\selectfont
\setlength{\tabcolsep}{0pt}
\begin{tabular*}{\textwidth}{@{\extracolsep\fill}>{ \raggedright\arraybackslash }p{0.17\textwidth}lcccc@{}}
\toprule
 & & \multicolumn{2}{c}{\textbf{Fixed screening capacity (20\%)}} & \multicolumn{2}{c}{\textbf{80\% sensitivity target}} \\
\cmidrule(lr){3-4}\cmidrule(lr){5-6}
\textbf{Setting (rural, prev.)} & \textbf{Group} & \textbf{Cases found} & \textbf{False alarms} & \textbf{Cases found} & \textbf{False alarms} \\
\midrule
US (observed) & rural & $+$157 ($+$72 to $+$246) & $+$653 ($+$249 to $+$1,093) & $-$48 ($-$87 to $-$18) & $-$493 ($-$891 to $-$193) \\
 & urban & $-$70 ($-$171 to $+$32) & $-$740 ($-$1,188 to $-$321) & $+$50 ($-$38 to $+$134) & $+$288 ($-$579 to $+$1,040) \\
 & all births & $+$87 ($-$13 to $+$191) & $-$87 ($-$191 to $+$12) & $+$3 ($-$114 to $+$102) & $-$205 ($-$1,452 to $+$735) \\
\midrule
World (42.7\%, 17.2\%) & rural & $+$337 ($+$137 to $+$545) & $+$948 ($+$329 to $+$1,574) & $-$175 ($-$305 to $-$63) & $-$1,300 ($-$2,097 to $-$521) \\
 & urban & $-$214 ($-$379 to $-$65) & $-$1,070 ($-$1,667 to $-$433) & $+$45 ($-$35 to $+$119) & $+$187 ($-$371 to $+$676) \\
 & all births & $+$123 ($-$54 to $+$299) & $-$123 ($-$299 to $+$55) & $-$130 ($-$317 to $+$29) & $-$1,113 ($-$2,394 to $-$11) \\
\midrule
World (42.7\%, 19.2\%) & rural & $+$369 ($+$151 to $+$593) & $+$927 ($+$323 to $+$1,485) & $-$195 ($-$339 to $-$70) & $-$1,266 ($-$2,044 to $-$507) \\
 & urban & $-$247 ($-$420 to $-$75) & $-$1,046 ($-$1,606 to $-$426) & $+$50 ($-$39 to $+$132) & $+$183 ($-$363 to $+$661) \\
 & all births & $+$122 ($-$67 to $+$319) & $-$120 ($-$318 to $+$67) & $-$145 ($-$353 to $+$32) & $-$1,083 ($-$2,334 to $-$8) \\
\midrule
Southern Asia (64.9\%, 22.3\%) & rural & $+$454 ($+$138 to $+$692) & $+$714 ($+$250 to $+$1,252) & $-$333 ($-$582 to $-$120) & $-$1,860 ($-$2,996 to $-$746) \\
 & urban & $-$289 ($-$483 to $-$131) & $-$878 ($-$1,303 to $-$445) & $+$34 ($-$27 to $+$91) & $+$109 ($-$215 to $+$394) \\
 & all births & $+$164 ($-$140 to $+$393) & $-$164 ($-$391 to $+$137) & $-$299 ($-$582 to $-$50) & $-$1,751 ($-$3,180 to $-$451) \\
\midrule
South-Eastern Asia (36.0\%, 13.5\%) & rural & $+$262 ($+$113 to $+$412) & $+$915 ($+$344 to $+$1,616) & $-$117 ($-$204 to $-$42) & $-$1,148 ($-$1,849 to $-$460) \\
 & urban & $-$143 ($-$288 to $-$33) & $-$1,035 ($-$1,736 to $-$427) & $+$40 ($-$31 to $+$105) & $+$217 ($-$431 to $+$786) \\
 & all births & $+$119 ($-$29 to $+$240) & $-$119 ($-$240 to $+$31) & $-$77 ($-$213 to $+$40) & $-$931 ($-$2,224 to $+$159) \\
\midrule
Southern Africa (36.4\%, 40.0\%) & rural & $+$716 ($+$312 to $+$1,118) & $+$559 ($+$227 to $+$970) & $-$350 ($-$609 to $-$126) & $-$765 ($-$1,252 to $-$304) \\
 & urban & $-$490 ($-$847 to $-$216) & $-$783 ($-$1,173 to $-$376) & $+$117 ($-$91 to $+$309) & $+$154 ($-$303 to $+$556) \\
 & all births & $+$226 ($-$51 to $+$483) & $-$224 ($-$484 to $+$54) & $-$234 ($-$637 to $+$115) & $-$611 ($-$1,514 to $+$142) \\
\midrule
South America (19.0\%, 21.7\%) & rural & $+$285 ($+$128 to $+$461) & $+$590 ($+$220 to $+$1,083) & $-$102 ($-$176 to $-$36) & $-$536 ($-$871 to $-$214) \\
 & urban & $-$128 ($-$332 to $+$24) & $-$746 ($-$1,236 to $-$331) & $+$83 ($-$64 to $+$220) & $+$248 ($-$494 to $+$900) \\
 & all births & $+$157 ($+$1 to $+$305) & $-$157 ($-$303 to $-$1) & $-$19 ($-$210 to $+$150) & $-$288 ($-$1,352 to $+$555) \\
\midrule
Low-income countries (65.7\%, 17.2\%) & rural & $+$317 ($+$82 to $+$542) & $+$731 ($+$313 to $+$1,280) & $-$260 ($-$455 to $-$93) & $-$2,017 ($-$3,244 to $-$810) \\
 & urban & $-$196 ($-$358 to $-$92) & $-$854 ($-$1,329 to $-$449) & $+$26 ($-$20 to $+$69) & $+$112 ($-$223 to $+$407) \\
 & all births & $+$121 ($-$112 to $+$334) & $-$123 ($-$334 to $+$116) & $-$234 ($-$455 to $-$41) & $-$1,905 ($-$3,427 to $-$504) \\
\botrule
\end{tabular*}
\vspace{4pt}\par\noindent{\footnotesize Change per 100,000 births (95\% CI; jurisdiction bootstrap). The simulator rebuilds each target population and re-runs the screening rule inside it. It reproduces the US estimates exactly at US composition, and it assumes that US score distributions within urban/rural $\times$ PPD strata carry over. The direction of the rural shift is the same at 10\% and 30\% capacity. \textit{Rural shares}: World Bank World Development Indicators, indicator SP.URB.TOTL.IN.ZS, based on UN World Urbanization Prospects (2023 values; East Asia \& Pacific and South Africa used as proxies for South-Eastern Asia and Southern Africa)~\citep{bib41,bib42}. \textit{Prevalence}: regional pooled estimates from \citet{bib21} and the global estimate from \citet{bib22}. \textit{Births}: US 3,596,017 (2023)~\citep{bib44}; world 132.1 million (2023)~\citep{bib45}.}
\end{table}

\begin{table}[htbp]
\caption{Cohort flow}\label{exttab6}%
\setlength{\tabcolsep}{0pt}
\begin{tabular*}{\textwidth}{@{\extracolsep\fill}lcc@{}}
\toprule
\textbf{Step} & \textbf{Remaining} & \textbf{Excluded} \\
\midrule
Phases 7--8 respondents & 390,320 & --- \\
Both outcome items answered & 380,689 & 9,631 \\
Core predictors complete & 356,626 & 24,063 \\
NCHS urban/rural known (1 or 2) & 350,723 & 5,903 \\
Jurisdiction present in PDFM (absent: AK, HI) & 332,970 & 17,753 \\
Urban/rural stratum present in PDFM & 332,970 & 0 \\
\botrule
\end{tabular*}
\vspace{4pt}\par\noindent{\footnotesize Alaska and Hawaii are absent from the PDFM v0 release.}
\end{table}

\end{document}